\pdfoutput=1
\documentclass[11pt]{article}

\usepackage[final]{acl}
\usepackage{comment}
\usepackage{times}
\usepackage{latexsym}
\usepackage{bbm}
\usepackage{booktabs}
\usepackage{amssymb} 
\usepackage{makecell}
\usepackage{multirow}   % Span cells across multiple rows
\usepackage{bm}         % Bold math symbols (\bm{...})
\usepackage{array}      % Better table column formatting (usually already included)
\usepackage[T1]{fontenc}
\usepackage[utf8]{inputenc}

\usepackage{microtype}

\usepackage{inconsolata}

\usepackage{graphicx}
\usepackage{algorithm}
\usepackage{algorithmic}
\usepackage{amsmath}
\usepackage{xurl}  % allows breaks at more characters in URLs
\usepackage{amsfonts}

\title{Audit Me If You Can: \\Query-Efficient Active Fairness Auditing of Black-Box LLMs}

 \author{
   \textbf{David Hartmann\textsuperscript{1,2}},
   \textbf{Lena Pohlmann\textsuperscript{1,2}},
   \textbf{Lelia Hanslik\textsuperscript{2}},\\
   \textbf{Noah Gießing\textsuperscript{3}},
   \textbf{Bettina Berendt\textsuperscript{1,2,4}},
   \textbf{Pieter Delobelle\textsuperscript{0,4, 5}}\\
   \textsuperscript{1}Weizenbaum Institut Berlin,
   \textsuperscript{2}Technische Universität Berlin,
   \textsuperscript{3}FIZ Karlsruhe,
   \textsuperscript{4}KU Leuven
   \textsuperscript{5}Pleias
}
\begin{document}
\maketitle
\begin{abstract}
Large Language Models (LLMs) exhibit systematic biases across demographic groups. Auditing is proposed as an accountability tool for black-box LLM applications, but suffers from resource-intensive query access. We conceptualise auditing as uncertainty estimation over a target fairness metric and introduce BAFA, the Bounded Active Fairness Auditor for query-efficient auditing of black-box LLMs. BAFA maintains a version space of surrogate models consistent with queried scores and computes uncertainty intervals for fairness metrics (e.g., $\Delta_{\mathrm{AUC}}$) via constrained empirical risk minimisation. Active query selection narrows these intervals to reduce estimation error. We evaluate BAFA on two standard fairness dataset case studies: \textsc{CivilComments} and \textsc{Bias-in-Bios}, comparing against stratified sampling, power sampling, and ablations. BAFA achieves target error thresholds with up to 41$\times$ fewer queries than stratified sampling (e.g., 144 vs 5,956 queries at $\varepsilon=0.02$ for \textsc{CivilComments}) for tight thresholds, demonstrates substantially better performance over time, and shows lower variance across runs. These results suggest that active sampling can reduce resources needed for independent fairness auditing with LLMs, supporting continuous model evaluations.

\end{abstract}
\section{Introduction}
\footnotetext[0]{Work done while at Aleph Alpha}
LLMs are increasingly deployed not only for generative tasks such as text completion, image synthesis, and video generation, but also for downstream decision-making tasks, including classification, scoring, and ranking. These systems are commonly offered via machine-learning-as-a-service (MLaaS) APIs and have substantial real-world impact, for example, in automated hate speech detection and candidate screening in hiring.

However, recent evaluations have shown that such applications exhibit systematic performance disparities across social groups. Commercial hate speech detection systems based on black-box LLMs have been found to underperform for LGBTQIA+ and people with disabilities~\cite{rottgerHateCheckFunctionalTests2021a, hartmannLostModerationHow2025}. Similarly, LLM-based CV and biography screening systems show biases with respect to disability status~\cite{glazkoIdentifyingImprovingDisability2024}, gender~\cite{wangJobFairFrameworkBenchmarking2024}, or educational background~\cite{isoEvaluatingBiasLLMs2025}.
\begin{figure}
    \centering
    \includegraphics[width=\linewidth]{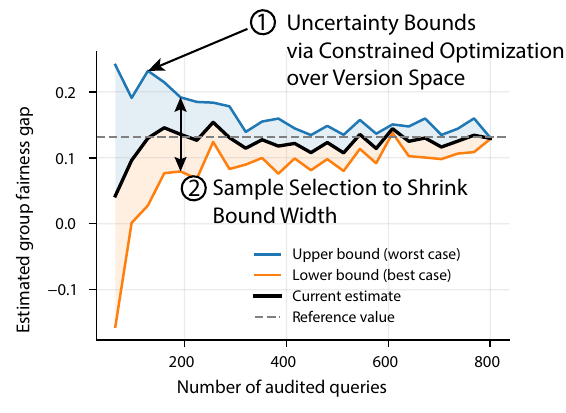}
    \caption{\textbf{Bounded Active Fairness Auditing (BAFA)}. Upper and lower bounds on the fairness metric converge as queries accumulate. BAFA aims to maximally shrink the uncertainty interval between bounds.}
    \label{fig:firstpage}
\end{figure}

To uncover such systemic risks in deployed systems, audits have been proposed as a key accountability mechanism~~\citep{raji2020auditing, birhaneAIAuditingBroken2024}. Independent black-box auditing is increasingly reflected in policy frameworks, including Appendix 3.5 of the EU Code of Practice on Generative AI, and it is widely discussed in governance and regulatory proposals~\cite{mokander2023auditing, raji_outsider_2022, hartmannAddressingRegulatoryGap2024a}.

\begin{figure*}
    \centering
    \includegraphics[width=0.95\linewidth]{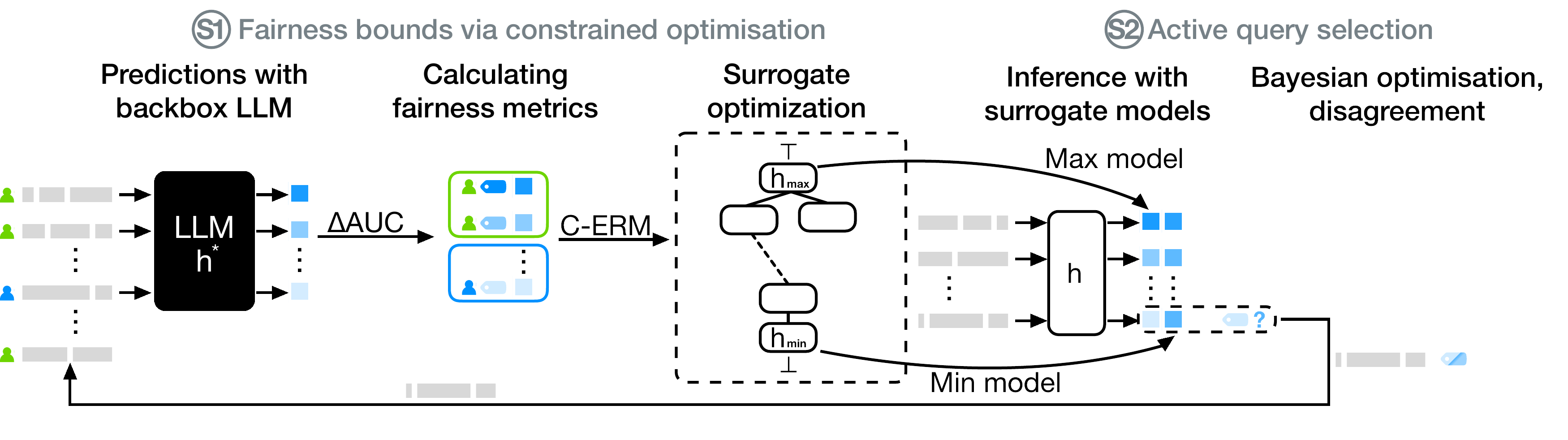}
    \caption{\textbf{BAFA Pipeline in more detail}. In every turn we sample $k$ samples. First, we query the black-box LLM with a stratified seed set from our dataset (1). Then, we calculate the estimated fairness measure (2) and do constraint optimisation with BERT surrogates (3) to get lower and upper fairness bounds. Based on the calculated scores for each $x\in D$ from the upper and lower models (4), BAFA selects queries (5) which shrink the distance between the lower and upper in high-disagreement regions, leading to faster and more stable convergence.}
    \label{fig:bafa}
\end{figure*}

In practice, however, conducting fairness audits of black-box LLMs remains challenging. Comprehensive audits typically require extensive amount of API queries, which are costly~\cite{hartmannLostModerationHow2025}, may raise privacy concerns~\cite{zaccourAccessDeniedMeaningful2025}, and can conflict with data minimisation obligations under the GDPR~\cite{rastegarpanah2021dataminimization}.

These challenges are especially pronounced in continuous auditing scenarios, where linguistic change and system updates necessitate repeated evaluations over time. A common workaround is the use of hand-crafted bias benchmarks or templates~\citep[\textit{e.g.,}~][]{rottgerHateCheckFunctionalTests2021a, nadeem2021stereoset, gehman2020realtoxicityprompts, nangia2020crows, rottger2022multilingual}. While useful for controlled testing, these approaches often lack ecological validity and provide limited reliability when auditing dynamic, context-sensitive tasks such as hate speech detection or biography scoring in real-world settings~\citep{delobelle2024metrics}.

Several works therefore argue for query-efficient, ecologically valid fairness auditing that operates under strict budget and black-box access constraints, enabling continuous evaluation by independent auditors~\cite{cenTransparencyAccountabilityBack2024, hartmannLostModerationHow2025}. While~\citet{TomZhang} propose an oracle-efficient method for auditing demographic parity, their approach does not scale to large hypothesis spaces such as LLMs and is incompatible with ranking-based fairness metrics commonly used in content moderation and hiring. Related work on query-efficient red teaming~\citep[\textit{e.g.,}~][]{lee2023query} actively surfaces harmful behaviours but cannot estimate specific fairness parameters in black-box settings. This gap motivates a query-efficient active fairness auditing method for black-box LLMs that supports ranking metrics.

\paragraph{Our approach.}
Bounded Active Fairness Auditing is introduced as a query-efficient auditing framework for black-box language models. As illustrated in Figure \ref{fig:firstpage}, BAFA conceptualises auditing as measuring uncertainty associated with a model's group fairness parameter, such as the group-wise ROC-AUC difference, within a specified query budget. This is achieved by optimising surrogate upper and lower bounds, colour-marked in one illustrative run. These represent an interval of plausible values for a fairness metric, based on the outputs obtained thus far. 

BAFA then actively selects queries that are expected to most effectively reduce uncertainty regarding the target fairness metric, thereby minimising the required query budget as demonstrated in the BAFA pipeline in Figure \ref{fig:bafa}. By focusing queries on fairness-critical regions of the input space, BAFA significantly reduces audit costs. Experimental results show that BAFA requires substantially fewer queries than baselines, performing better over time and achieving lower variance, evaluated in two practical auditing scenarios.

The contributions of this work are threefold: (1) \textbf{methodological:} we present an active fairness auditing method for black-box LLMs that works with threshold-invariant ranking metrics -- to the best of our knowledge, 
the first active learning approach for black-box LLM 
fairness auditing,
(2) \textbf{practical:} we introduce a query-efficient framework suitable for independent audits under limited access, budget, and regulatory constraints, and (3) \textbf{empirical:} we compare several samplings methods for auditing and substantial query-efficiency gains over three baseline sampling approaches in two realistic LLM auditing case studies.

\section{Related Work}

\paragraph{Fairness evaluation of language models.}
LLMs exhibit systematic demographic biases~\cite{blodgettLanguageTechnologyPower2020a}, as demonstrated by hate speech detection~\cite{sap-etal-2019-risk} and CV scoring~\cite{glazkoIdentifyingImprovingDisability2024}. Prior work has relied heavily on template-based benchmarks such as HateCheck~\cite{rottgerHateCheckFunctionalTests2021a}, StereoSet~\cite{nadeem2021stereoset}, and RealToxicityPrompts~\cite{gehman2020realtoxicityprompts}, which enable controlled comparisons but suffer from limited construct validity and weak alignment with real-world use~\cite{goldfarb-tarrant2021intrinsic}. As a result, these benchmarks provide static snapshot evaluations that are ill-suited for auditing deployed systems over time~\cite{tonneau2025hateday}. Critically,~\citet{blodgett-etal-2021-stereotyping} argue that benchmark-driven evaluations often conflate distinct notions of bias and obscure concrete group-level harms, motivating auditing approaches grounded in real-world data and explicit fairness metrics.

\paragraph{Black-box auditing and red teaming.}
Under black-box access and beyond benchmark-driven evaluation, two dominant evaluation paradigms have emerged: red teaming and auditing. Red teaming seeks to uncover worst-case or unsafe behaviours through adversarial querying, providing evidence of failure modes without estimating their prevalence~\cite{perez2022red}. In contrast, black-box auditing aims to estimate well-defined system properties, such as fairness, via systematic black-box queries, potentially conducted by independent external stakeholders~\cite{raji_outsider_2022,mokander2023auditing}. However, comprehensive audits are often infeasible in practice due to high query costs, rate limits, and legal constraints such as GDPR data minimisation~~\cite{rastegarpanah2021dataminimization,zaccourAccessDeniedMeaningful2025}. These constraints motivate query-efficient auditing methods that can provide reliable estimates within strict budgets.
\paragraph{Query-efficient and active auditing.}
Query-efficient auditing seeks to estimate a fairness measure of a black-box using as few queries as possible.
Existing work has focused on sample size reduction via rigorous passive sampling approaches. For example,~\citet{singh2023brieftutorialsamplesize} derive closed-form requirements for detecting fairness violations under power sampling, but do not consider adaptive query selection. While active learning reduces label complexity by selecting informative examples~\cite{settles.tr09}, most approaches optimise predictive performance rather than group-level fairness estimation. Recent frameworks cover related areas, like online monitoring with confidence sequences~\cite{Maneriker}, Fourier fairness coefficients for discretised inputs, ~\citet{ajarra2024active}, and Bayesian Optimisation (BO) for red teaming~\cite{lee2023query}. However, they do not support group fairness auditing when statistical uncertainties are present.
Active fairness auditing using constrained empirical risk minimisation (C-ERM)~\cite{TomZhang} offers strong guarantees for threshold-based metrics. However, it depends on optimisation surrogates that are not practical for modern LLMs, since these surrogates must closely mimic the black-box model. Most active auditing frameworks also focus on threshold-dependent classification metrics, even though many commercial models produce continuous scores. Both~\citet{TomZhang} and~\citet{singh2023brieftutorialsamplesize} have called for extensions to these metrics. For such systems, threshold-invariant measures like group-wise ROC AUC difference are more suitable, as they capture disparities across all possible decision thresholds~\cite{borkan2019limitationspinnedaucmeasuring, borkan, gallegos-etal-2024-bias}.

\section{Bounded Active Fairness Auditing}
\paragraph{Black-box Audit Setup.}
We audit a black-box model $h^\star$ that assigns scores to inputs (e.g., toxicity scores for comments, confidence scores for occupation predictions). Given labeled data with ground-truth labels $y_i$ and protected group attributes $g_i \in \{0,1\}$, our goal is to estimate the ranking fairness gap between two demographic groups:
\[
\Delta_{\text{AUC}}(h^\star) = \text{AUC}_{g=0}(h^\star) - \text{AUC}_{g=1}(h^\star),
\]
where $\text{AUC}_{g}$ measures how well the model ranks positive examples above negative examples for group $g$. Given a query budget $T$ (e.g., 1000 API calls), we seek an estimator $\widehat{\Delta}_{\text{AUC}}$ that is $\epsilon$-accurate (e.g., within $\pm 0.02$ of the true disparity) while minimising the number of queries $q \leq T$ needed. We assume access to ground-truth labels and group attributes for evaluation, but only black-box access to the model itself (complete mathematical formulation can be found in App. \ref{app:formal_setup}).

\paragraph{Algorithm Overview.}
Figure~\ref{fig:bafa} summarises our method, Bounded Active Fairness Auditing (BAFA)\footnote{\url{https://github.com/dawiethart/AuditMeIfYouCan-Active-Auditing}}. 
Starting from a stratified seed set, BAFA iteratively (\textbf{S1}) computes upper and lower fairness bounds via constrained optimisation, and (\textbf{S2}) selects new queries that are expected to
maximally reduce the bound width and thus, uncertainty in the group fairness measure. 
\begin{algorithm}[t]
\caption{Bounded Active Fairness Auditing (BAFA)}
\label{alg:bafa_short}
\begin{algorithmic}[1]
\REQUIRE Audit pool $\mathcal{U}$, black-box $h^\star$, budget $T$, 
         tolerance $\lambda$, batch size $k$, threshold $\epsilon$, fairness measure $\mu$, hypothesis space $\mathcal{H}$
\STATE $S_0 \leftarrow$ stratified seed set; query $h^\star$ to attach scores $\{s_i^\star\}$
\FOR{$t = 0, 1, 2, \ldots, T$}
    \STATE \textbf{// S1: Certificate}
    \STATE Solve $h_{\max}^t \leftarrow \arg\max_{h \in \mathcal{H}_\lambda(S_t)} \mu(h)$
    \STATE Solve $h_{\min}^t \leftarrow \arg\min_{h \in \mathcal{H}_\lambda(S_t)} \mu(h)$
    \STATE $[\mu_{\min}^t, \mu_{\max}^t] \leftarrow [\mu(h_{\min}^t),\, \mu(h_{\max}^t)]$
    \IF{$(\mu_{\max}^t - \mu_{\min}^t)/2 \leq \epsilon$}
        \RETURN $\hat{\mu}_t = (\mu_{\min}^t + \mu_{\max}^t)/2$
    \ENDIF
    \STATE \textbf{// S2: Active query selection}
    \FOR{each $x \in \mathcal{U} \setminus S_t$}
        \STATE $\mathrm{dis}_t(x) \leftarrow |h_{\max}^t(x) - h_{\min}^t(x)|$
    \ENDFOR
    \STATE $Q_t \leftarrow \mathrm{top}\text{-}k\text{ candidates by } \mathrm{dis}_t(x)$
    \STATE Query $h^\star$ on $Q_t$; update $S_{t+1} \leftarrow S_t \cup Q_t$
\ENDFOR
\STATE \textit{// Implementation details: App.~\ref{app:impl_details_full}}
\end{algorithmic}
\end{algorithm}

\begin{description}
\item[S1:] \textbf{Fairness bounds via constrained optimisation.} BAFA quantifies uncertainty in fairness
by maintaining a set of surrogate hypotheses
that are consistent with the black-box model
on the queried set $S$. Specifically, BAFA maintains a $\lambda$-approximate version space $\mathcal{H}_\lambda(S_t) = \{h \in \mathcal{H} : |h(x_i) - s_i^\star| \leq \lambda,\  \forall (x_i, s_i^\star) \in S_t\}$ of BERT-based surrogates~\cite{devlin2019bert} consistent with observed  black-box scores. In each round, we solve two constrained problems via  Cooper~\cite{gallegoPosada2025cooper} where one is maximising, and one is minimising  $\Delta_\mathrm{AUC}$ over $\mathcal{H}_\lambda(S_t)$. This yields extremal  hypotheses $h_{\max}^t, h_{\min}^t$ and the certificate interval  $[\mu_{\min}^t, \mu_{\max}^t]$. As ROC-AUC is non-differentiable, we optimise a sigmoid pairwise ranking  surrogate~\citep{pmlr-v30-Agarwal13}. As $|S_t|$ grows,  $\mathcal{H}_\lambda(S_t)$ shrinks monotonically and the interval narrows  (see App.~\ref{app:formal_setup}).

\item[S2:] \textbf{Active query selection.} To reduce required queries, BAFA selects inputs expected to shrink the fairness uncertainty interval the most. We operationalise this by scoring each unqueried candidate via the disagreement of the two extremal surrogates from S1: \[ \mathrm{dis}_t(x) = \bigl|h_{\max}^t(x) - h_{\min}^t(x)\bigr|, \] and querying the top-$k$ highest-scoring inputs per round.
  We adopt top-$k$ rather than the $\varepsilon$-driven disagreement loop of~\citet{TomZhang}, which iterates until the surrogate itself falls below a diameter threshold. Such schemes require the surrogate to reliably mimic the black box. This condition holds only after roughly 500--750 queries in our setting (App.~\ref{app:surrogate}). Top-$k$ instead uses the surrogate \emph{only} to rank candidates by bound disagreement, not as a black-box proxy. This is a strictly weaker requirement as it demands consistent \emph{ordering} of candidates rather than accurate \emph{score replication}, and remains effective even under partial surrogate--black-box agreement of scores (App.~\ref{app:surrogate}). 
\end{description}
\begin{table*}[ht!]
\centering
\resizebox{\textwidth}{!}{%
\renewcommand{\arraystretch}{1.15}
\begin{tabular}{ll||cc|cccc}
\toprule
\textbf{Case Study} & $\bm{\varepsilon}$
& \makecell{\textbf{BAFA}\\\textbf{(Dis.)}}
& \makecell{\textbf{BAFA}\\\textbf{(BO)}}
& \makecell{\textbf{C-ERM}\\\textbf{(abl.)}}
& \makecell{\textbf{BO only}\\\textbf{(abl.)}}
& \makecell{\textbf{Power}\\\textbf{(base.)}}
& \makecell{\textbf{Stratified}\\\textbf{(base.)}} \\
\midrule
\multicolumn{8}{l}{\textit{Queries to $\varepsilon$ (mean [95\% CI]) $\downarrow$}} \\
\multirow{2}{*}{\textsc{CivilComments}}
& 0.02 & \textbf{144 [98, 190]}  & 256 [180, 332]
       & 457 [320, 594]           & 1{,}204 [864, 1{,}544]
       & 8{,}436 [7{,}876, 8{,}836] & 5{,}956 [1{,}268, 7{,}652] \\
& 0.05 & \textbf{80 [58, 102]}   & 132 [96, 168]
       & 137 [102, 172]           & 356 [240, 472]
       & 932 [484, 2{,}756]       & 452 [164, 676] \\
\multirow{2}{*}{\textsc{Bias-in-Bios}}
& 0.02 & \textbf{340 [248, 432]} & 356 [268, 444]
       & 512 [380, 644]           & 772 [580, 964]
       & 5{,}396 [516, 5{,}988]   & 1{,}748 [564, 3{,}060] \\
& 0.05 & 148 [108, 188]          & 180 [136, 224]
       & 210 [158, 262]           & \textbf{100 [72, 128]}
       & 356 [4, 372]             & 212 [4, 324] \\
\midrule
\multicolumn{8}{l}{\textit{Mean AUEC for first 1k queries $\downarrow$}} \\
\textsc{CivilComments} &
& \textbf{0.019} & 0.022 & 0.030 & 0.060 & 0.093 & 0.066 \\
\textsc{Bias-in-Bios} &
& \textbf{0.025} & 0.029 & 0.042 & 0.035 & 0.045 & 0.042 \\
\midrule
\multicolumn{8}{l}{\textit{Error at 250 queries (mean [95\% CI]) $\downarrow$}} \\
\textsc{CivilComments} &
& \textbf{0.020 [0.015, 0.026]} & 0.021 [0.013, 0.030]
& 0.030 [0.015, 0.046]          & 0.096 [0.062, 0.131]
& 0.108 [0.080, 0.135]          & 0.064 [0.046, 0.083] \\
\textsc{Bias-in-Bios} &
& 0.022 [0.017, 0.027] & \textbf{0.022 [0.017, 0.026]}
& 0.024 [0.014, 0.033]  & 0.023 [0.014, 0.033]
& 0.065 [0.045, 0.085]  & 0.043 [0.028, 0.058] \\
\bottomrule
\end{tabular}%
}
\caption{\textbf{BAFA substantially reduces query costs in both case studies while 
beating baselines in over-time performance and stability across 20 seeds.}
We report (i) \emph{convergence query-efficiency} as queries required until the mean 
error curve falls under $\varepsilon$, with bootstrapped 95\% CIs across 20 seeds; 
(ii) \emph{over-time performance} via $\mathrm{AUEC}$ over the first 1k queries; and 
(iii) \emph{mid-budget error} at 250 queries with bootstrapped 95\% CIs.
Dis.\ = disagreement-based selection. Wide CIs for baselines at strict thresholds 
reflect high run-to-run variability, further supporting the case for active auditing.}
\label{tab:main_results}
\end{table*}

Two disagreement-based scoring rules are used and evaluated that do not require an accurate surrogate for query selection. First, \emph{Bound-disagreement sampling} prioritises candidates where the current upper- and lower-bound models of S1 disagree most on AUC-relevant pairwise rankings.
Second, as a comparison to \emph{Bound-disagreement sampling}, we test \emph{Bayesian optimisation} which searches over acquisition features -- including bound disagreement, LoRA-surrogate diversity, and surrogate--black-box disagreement -- as a proxy for uncertainty. Inspired by~\citet{lee2023query}, this should balance between exploitation of high-impact regions and exploration for text diversity. For both strategies, we apply distributional regularisation using empirical subgroup and label marginals to mitigate selection-induced bias in fairness estimation (Details, see App. \ref{app:impl_details_full}).
\section{Experimental Setup}
BAFA is evaluated in two black-box LLM deployments under realistic audit constraints: (A) hate speech detection and (B) profession estimation from biographies. The two case studies are deliberately complementary: Case Study A uses a locally-hosted classifier with a known, synthetically injected disparity ($\mu_{\Delta\mathrm{AUC}} \approx 0.14$)  and an audit pool of $\sim$50k comments, providing a controlled setting with a severe fairness violation; Case Study B uses a commercial black-box (GPT-4.1-mini) with a non-injected disparity ($\mu_{\Delta\mathrm{AUC}} \approx 0.02$--$0.045$) and scores over $\sim$50k biographies, testing BAFA under architectural mismatch between surrogate and target. Together, they span the range of realistic independent audit settings with two high-stakes scenarios that should be independently evaluated.

In both case studies, the auditor has access only to model inputs and outputs and seeks to estimate group-level ROC AUC disparities under a fixed query budget. All strategies are evaluated using a common protocol with identical budgets and batch sizes, and the results are averaged across 20 random seeds. At each audit round, a batch of inputs is selected for black-box querying with each strategy, and the fairness estimate is updated.

In Table \ref{tab:main_results}, we report \emph{convergence query-efficiency} as the number of black-box queries needed until the \emph{mean} absolute error across seeds first falls below a target threshold $\varepsilon \in \{0.02, 0.05\}$. Additionally, we report \emph{over-time performance} via the area under the error curve ($\mathrm{AUEC}$) over the first $1000$ queries (analogous to AUC), and quantify \emph{stability} by the mean error and standard deviation across seeds at fixed budgets.
We compare BAFA against stratified and power sampling (calculated for $\Delta_\mathrm{AUC}$ from~\citet{singh2023brieftutorialsamplesize}) as baselines, constrained optimisation (as in~\citet{TomZhang} with a stratified sample), and BO without active querying as ablations. These baselines and ablations allow us to disentangle the effects of active selection and constrained optimisation. Complete evaluation metric details (App.~\ref{app:metrics}), baseline and ablations definitions (App.~\ref{app:baselines}) as well as implementation details (App.~\ref{app:exp_details}) are provided in Appendix. A sensitivity analysis covering $\lambda$, batch size $k$, 
regularisation weight $\alpha$, and C-ERM optimization epochs 
is provided in App.~\ref{app:exp_details} 
(Tables~\ref{tab:hyperparameter_ablation}--\ref{tab:batch_sizes_with_error}).

\section{Results}

Table~\ref{tab:main_results} summarises the query efficiency and estimation performance of different auditing strategies over 20 random seeds in both case studies. 
\subsection{Case Study A: Auditing Hate Speech Detection}

Our first case study audits group-based performance disparities in hate speech detection using real-world, identity-labelled data. We use the \textsc{CivilComments} dataset~\cite{borkan}, which contains user-generated public comments on English-language news sites, annotated for toxicity and multiple identity targets. We focus on eight target groups commonly studied in prior work (e.g., gender, religion, sexual orientation) and evaluate disparities between dominant and marginalised groups (For details see App. \ref{app:exp_details}).

As the audited system, we construct a controlled but highly biased black-box model by fine-tuning HateBERT \cite{caselli-etal-2021-hatebert} on the SBIC dataset \cite{sap_social_2020}, systematically flipping labels for comments targeting marginalised groups ($\mu_{\Delta AUC} \approx 0.14$ for each group pair). This synthetic setup provides a known and severe fairness violation, allowing us to assess whether active auditing can reliably detect disparities under limited query budgets.

\begin{figure}[t!]
  \centering
  \includegraphics[width=\columnwidth]{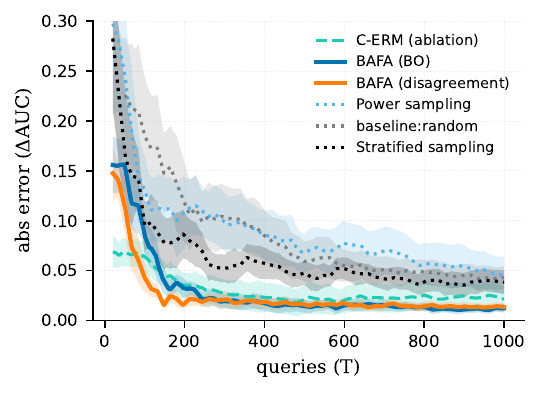}
  \caption{\textbf{Active auditing methods perform more query-efficient and stable over 20 \textsc{Civil Comments} seeds.} BAFA methods (solid) converge significantly faster than baseline sampling strategies (dotted). Shaded areas indicate 95\% confidence intervals across seeds and demonstrate that BAFA methods show substantially reduced variance compared to baseline methods.}
  \label{fig:plot_time_1}
\end{figure}

\paragraph{Query efficiency to target threshold.} Across both thresholds, active auditing strategies require substantially fewer queries than passive baselines to reach a given accuracy. For $\varepsilon = 0.02$, BAFA with disagreement and BAFA with BO reach the target error within 144--256 queries on average, whereas stratified and power sampling require several thousand queries, ablations around 2--8$\times$ more. Disagreement-based sampling is approximately 41$\times$ faster than stratified sampling for $\varepsilon=0.02$. The result is a bit less pronounced for the less stricter threshold $\varepsilon = 0.05$, where both BAFA approaches reduce the mean of queries needed around three to five times ($5.7$ for disagreement and $3.4$ for BO) in relation to stratified sampling. 
\paragraph{Over-time estimation accuracy.}
Presented in Figure~\ref{fig:plot_time_1} and by mean $\mathrm{AUEC}$, our active methods also perform substantially better (error-reduction around 3-4 times) than baselines in terms of over-time performance. Over the first 1,000 queries, BAFA with disagreement achieves the lowest mean AUEC on \textsc{CivilComments}, followed closely by BAFA with BO. In contrast, stratified and power sampling accumulate substantially higher error over time due to slow early progress, whereas after 1,000 queries, $\mathrm{AUEC}$ is similar across all approaches. Interestingly, C-ERM already outperforms both baselines (see Figure~\ref{fig:plot_time_1}) and BO without active sampling, but its performance is still below that of the BAFA variants.
\paragraph{Mid-budget accuracy and stability.}
At a mid-range budget of 250 queries, BAFA with disagreement achieves the lowest mean estimation error on \textsc{CivilComments} with reduced variance across seeds, which are also visible in CI-bands in Figure~\ref{fig:plot_time_1}, making it the most reliable estimator at fixed budgets. BAFA with BO is close in mean error but exhibits higher variance at this point, representing a trade-off between early exploration and overall stability, although mean AUEC are comparable across both BAFA approaches. Baselines demonstrate substantially larger error bands at 250 queries and show large run-to-run variability. 
\subsection{Case Study B: Auditing Black-Box CV Scoring LMs}
Our second case study examines fairness in automated hiring scenarios by auditing a black-box language model used for occupation inference. We use the \textsc{Bias-in-Bios} dataset~\cite{dearteaga2019bias}, which contains short biographies annotated with ground-truth occupations and binary gender labels. We use GPT-4.1-mini  as a black-box scorer via a deterministic prompt that maps biographies to (i) a predicted occupation from a predefined label set and (ii) a confidence score in $[0,100]$.

A small, disjoint subset of biographies is used as few-shot examples to stabilise model behaviour; the remaining biographies form the audit dataset. For each occupation, we define a binary classification task (target occupation vs.\ all others), using the model’s confidence score as a ranking signal. Group-wise ROC-AUCs are computed separately for male and female biographies, and fairness is again measured via $\Delta_{\text{AUC}}$ ($\mu_{\Delta AUC} \approx 0.02-0.045$) (Details App. \ref{app:exp_details}).

This case study complements content moderation by testing our method in a distinct, potentially biased domain with different data distributions and a commercial black-box model that is qualitatively different from our surrogate in both architecture and scale. 
One open question is whether BAFA performs better even for such substantially larger black-box models, since our C-ERM step uses a comparatively small BERT surrogate to reduce the \emph{fairness-metric version space} induced by queried scores rather than the black box’s full parameter space; we address this question empirically in this case study.
\begin{figure}[t]
  \centering
  \includegraphics[width=\columnwidth]{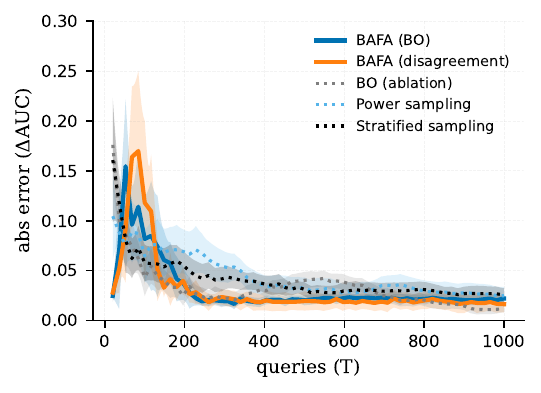}
  \caption{\textbf{Active auditing methods perform even with large parameter spaces with \textsc{GPT-4.1-mini} as black-box.} Similarly, to Fig. \ref{fig:plot_time_1}, BAFA methods converge significantly faster than baseline sampling strategies and show substantially reduced variance compared to baseline methods. However, a much bigger variance and worse performance are visible for the first 100-120 queries, probably related to the model mismatch.}
  \label{fig:plot_time_2}
\end{figure}
\paragraph{Query efficiency to target threshold.}
Again, on \textsc{Bias-in-Bios}, active auditing converges faster than baselines when it comes to the strict accuracy thresholds. For $\varepsilon = 0.02$, both BAFA variants reach the target within around 340--356 queries on average, with disagreement performing slightly better than BO. Stratified sampling requires 1,748 queries and power sampling more than 5,300 queries, corresponding to roughly a $5\times$ and $16\times$ reduction, respectively. At the looser threshold $\varepsilon = 0.05$, BAFA's gains are smaller as it reaches the target within 148--180 queries, while baselines require 212 and 356 queries. Interestingly, for this threshold and case study, the ablation BO outperforms BAFA in convergence, although it needs about $2.3\times$ more queries for $\epsilon=0.02$.

\paragraph{Over-time estimation accuracy and stability.}
Consistent with Case Study A, active methods achieve lower error throughout the audit process. BAFA with disagreement yields the lowest AUEC over the first 1,000 queries, indicating faster uncertainty reduction across rounds, while BAFA with BO performs comparably but with slightly higher cumulative error early on. However, we acknowledge that the difference to baselines is less pronounced than in case study A, and Figure \ref{fig:plot_time_2} demonstrates that, although BAFA converges faster and is more stable after 100--120 queries, it shows more variance and larger error than baselines in the first 100--120 queries. At 250 queries, however, both BAFA variants achieve lower estimation error and higher stability than baselines.

\section{Discussion}
\paragraph{Active auditing can reduce query budget by a significant amount compared to baselines.}
Empirically, BAFA with both approaches reaches strict and loose targets with substantially fewer queries than baselines, maintains lower error throughout the audit, and is more stable at moderate budgets, especially on \textsc{CivilComments}, where baseline variability is high. 
Overall, the results indicate that BAFA is not just a query-efficient convergence algorithm, but a practical approach for producing more accurate and reproducible $\Delta\mathrm{AUC}$ estimates under access and resource constraints.
Furthermore, we observe a consistent trade-off between query selection methods: disagreement prioritises fast early interval shrinkage, while BO tends to achieve better mid-budget accuracy. Both approaches, however, seem to work similarly well, although our hypothesis was that BO would outperform simple disagreement. BAFA-BO, however, produces more reliable bound widths with high correlation to the absolute error.

\begin{figure}[t]
  \centering
  \includegraphics[width=\columnwidth]{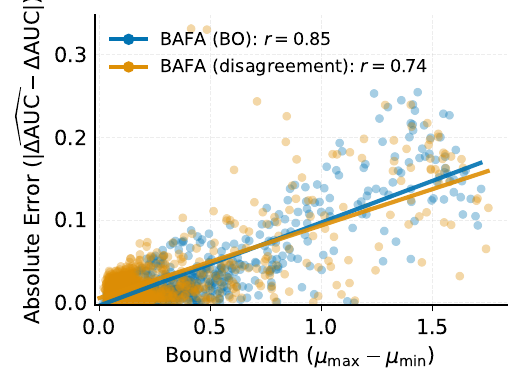}
  \caption{Relationship between BAFA’s uncertainty bound width and the true absolute estimation error of $\Delta$AUC in first 1k queries for \textsc{Civil Comments}. Each point corresponds to an audit during active querying. Bound width strongly correlates with actual error for both BO and disagreement-based selection.}
  \label{fig:fig5}
\end{figure}
\paragraph{Uncertainty calibration of BAFA.}
This view of auditing follows~\citet{TomZhang} and treats uncertainty as a version space quantity by computing an uncertainty interval $[\mu_{\min},\mu_{\max}]$ by solving two constrained optimisation problems that minimise and maximise the target metric over the version space. The resulting width has a direct operational meaning as it upper-bounds how much the estimated $\Delta \mathrm{AUC}$ could change under any hypothesis still compatible with the observed queries, and it shrinks as additional queries eliminate hypotheses from the version space. While a fully Bayesian approach would instead report credible intervals from a posterior, the version space formulation is computationally tractable for auditing large black-box LLMs with small surrogates. Empirically, Figure~\ref{fig:fig5} shows that bound width is strongly correlated with the true absolute estimation error, and that the ground-truth metric lies within BAFA’s interval in over 95\% of queries on \textsc{Civil Comments} (99.9\% for BAFA-BO and 95.4\% for disagreement). On Bias-in-Bios, coverage is lower due to surrogate--target mismatch, but the average bound violation remains below our strict tolerance $\varepsilon=0.02$, so width remains a useful proxy for uncertainty about the group-level fairness metric (Appendix~\ref{app:calibration}). 

\paragraph{Scaling to black-box LLMs with small surrogates.}
An obvious question in this setting is whether BAFA performs well even when the audited system is a large black-box LLM like GPT-4.1-mini, while using a much smaller surrogate such as BERT for constraint optimisation.
However, passive baselines achieve lower AUEC at very small budgets in Case Study~B because the underlying disparity ($\mu_{\Delta \mathrm{ROC\ AUC}} \approx 0.02$--$0.045$) is smaller than in Case Study~A, resulting in lower-variance $\Delta\mathrm{AUC}$ estimates at small sample sizes.
More importantly, beyond this initial phase, both BAFA variants outperform the baselines in convergence rate (reaching $\epsilon = 0.02$ with $\approx 5-41\times$ fewer queries than stratified sampling for case study A ($\approx41\times$) and B ($\approx5\times$), and by 250 queries they exhibit reduced variance and achieve AUEC over the first 1,000 queries that is approximately 60\% of the baseline AUEC, despite the larger early mean error.
In practice, BAFA reduces total AUEC and reaches target precision $\epsilon$ with fewer queries in this regime, and replacing the surrogate with DistilBERT increases AUEC by less than 5\% (for 3 seeds).
We interpret this as consistent with the version space view of~\cite{TomZhang} in that BAFA need not match the black box in parameter space, but must fit queried scores well enough that the constrained optimisation remains feasible and yields a non-trivial interval for $\Delta\mathrm{AUC}$ that shrinks as more informative queries are added. Nevertheless, when the surrogate and audited model are architecturally mismatched, the resulting intervals should be treated as an operational proxy rather than a coverage guarantee.
\paragraph{From failure discovery to quantified uncertainty.}
We treat auditing as \emph{uncertainty reduction over a target  metric}: given a query budget, the goal is not to find the worst  failure but to narrow the interval of plausible values for a  population-level property such as $\Delta\mathrm{AUC}$, until the  remaining uncertainty is small enough to support a defensible claim.  This contrasts with red teaming, which takes inspiration from  Bayesian sequential black-box testing~\cite{lee2023query} and  typically aims to surface as many failures as possible within a  fixed budget~\cite{feffer}, without estimating prevalence or  magnitude.
This also clarifies how our approach relates to hypothesis-testing in auditing:~\citet{cenTransparencyAccountabilityBack2024} argue that audits can be framed as hypothesis tests, which is useful for binary compliance decisions under a legal standard. Yet under limited budgets and potential distribution shift, conclusions become sensitive to the chosen threshold and prior assumptions \cite{juarez2022blackbox}. Reporting calibrated uncertainty about the audited quantity is therefore often more informative than a pass/fail certificate, and hypothesis tests can be treated as a downstream decision step, e.g., declaring non-compliance only if the entire uncertainty interval lies above a regulatory standard.

\paragraph{Implications for independent evaluation and continuous monitoring.}
This framework can support independent evaluators such as NGOs, journalists, and academic auditors in identifying downstream harms under constrained access. When black-box queries are costly, a smaller budget makes it feasible to audit more groups, domains, and languages, and to test targeted hypotheses about where harms may occur (e.g., subgroup-specific false positives that drive unfair moderation). More broadly, the results support continuous monitoring. Instead of running just one benchmark, an auditor can regularly check for disparities, e.g., after model updates, policy changes, or language evolutions, as was called for in hate speech moderation by~\citet{tonneau2025hateday} and~\citet{hartmannLostModerationHow2025}. One possible direction for future work is to see auditing as a process of information gain over time, where each new label updates our understanding of fairness and possible distribution shifts. This would help make better use of past audit data and allow for more flexible monitoring.

\paragraph{Understanding contextual implications of input query selection.}
Active auditing has an additional benefit, namely, that the sequence and composition of queried examples indicate which inputs are selected as most informative under its constraints, offering a potential form of interpretability (as in~\cite{activeLearningInterpret}). BAFA-BO builds on this by using a LoRA surrogate together with a query-diversity signal (as in~\cite{lee2023query}). This combination can make it even more interpretable for understanding selection patterns. Future work should build on this to characterise which regions of the input space different selection rules emphasise, for instance, borderline cases, specific linguistic patterns (e.g., AAE or counterspeech~\cite{sap-etal-2019-risk}), identity tokens or particular subpopulations. Such analyses could guide further qualitative investigation and stakeholder review of the sampled content.

\paragraph{Generalisability beyond ROC AUC difference.}
Lastly, while we instantiate our framework for $\Delta\mathrm{AUC}$, the broader idea is that active, query-efficient auditing can be applied whenever a black-box system exposes a reliable signal that can be turned into a scorable objective (differentiable or well-approximated by a smooth surrogate), enabling optimisation and uncertainty-aware selection. This covers other group metrics (e.g., TPR/FPR gaps at fixed thresholds, equalised odds, see~\citet{gallegos-etal-2024-bias}) and extends to performance~\cite{ribeiro-etal-2020-beyond}, privacy audits~\cite{staufer2025llms_forget} and robustness and safety audits~\cite{rauba2025statistical}, as well as benchmarking~\cite{liang2023helm}.

\section{Conclusion}

We presented BAFA, a query-efficient framework for auditing group fairness of black-box language models under realistic access and budget constraints. Across two auditing scenarios -- hate speech detection and profession inference -- BAFA consistently reduced the number of required queries by one order of magnitude compared to sampling baselines and ablations, while achieving lower estimation error and improved stability at moderate budgets. Conceptually, our results support viewing auditing as uncertainty estimation over a target metric rather than failure discovery or one-shot benchmarking. While BAFA does not resolve downstream harms or replace qualitative evaluation, it provides a practical measurement tool for making independent fairness audits with limited access more feasible, interpretable, and precise for black-box LLMs.

\section*{Limitations}
\paragraph{Surrogate model choice and the computational-precision trade-off}
We chose BERT-base as our surrogate model to keep computational costs low, which is important for independent auditors like civil society groups, journalists, and academic researchers who often have limited resources. There is a trade-off, though: the method works best when the surrogate model is similar to the black-box system being audited (though our ablations found only marginal differences when switching to DistillBERT). If auditors know the system’s architecture and have more resources, they can use a larger or better-matched surrogate, such as GPT-2 for auditing GPT-3, or RoBERTa-large for more complex tasks. Future work should especially try out GPT-2 or GPT-3 for the GPT-4.1-mini audit as architectures are the same and, thus, could lead to more accurate results and faster convergence. However, such experiments are out of scope for this work due to the focus on independent audits. In our experiments for Case Study B, we show that BAFA still performs very well even when the surrogate and target architectures do not match exactly.

\paragraph{Computational and resource intensity.}
A key limitation is that our end-to-end pipeline is resource intensive as it requires repeated optimisation steps within the loop. This is costly in wall-clock time and GPU usage, especially when scaling to many seeds, many groups, or frequent monitoring (see Appendix section \ref{app:comp_costs} for a detailed analysis of computational resources needed). This directly conflicts with our motivating goal of enabling resource-efficient auditing for independent evaluators. However, we think that substantial speedups are likely feasible. Promising directions include engineering improvements (e.g., caching/more efficient data pipelines), algorithmic warm-starting across rounds, more efficient batching strategies, and hybrid protocols that switch to simpler sampling once the interval is already narrow but a careful study of these system-level trade-offs is unfortunately out of scope for this work.
\paragraph{From a research prototype to an auditor-facing tool.}
While BAFA demonstrates the feasibility of query-efficient, uncertainty-aware auditing in controlled experimental settings, it is not yet a finished tool that can be readily deployed by independent auditors in practice. Turning BAFA into a practical auditing tool would therefore require integrating the needs and requirements of stakeholders and users, including support for multiple evaluation metrics (fairness-related or otherwise), transparent uncertainty reporting, and simple mechanisms for updating datasets and managing query budgets. A promising direction is the development of human-centered interfaces that allow auditors to configure audits through intuitive interactions (e.g., selecting metrics, uploading or modifying datasets, and issuing queries via clicks or drag-and-drop with uncertainty visualization). We see BAFA as a methodological building block toward such systems, but significant design, engineering, and participatory work remains to translate it into a robust and usable auditing infrastructure.

\paragraph{From metric gaps to downstream harms and the limits of ``certificates''.} 

Finally, fairness metrics (including bounded disparity estimates) are only proxies for real-world harm. Connecting a measured gap to downstream impacts requires context interpretations: whom the system affects, how it is used, and what policies and incentives  shape outcomes~\cite{blodgettLanguageTechnologyPower2020a}. In many cases, quantitative disparity estimates alone will not surface the most important harms~\cite{raji2021everything}. We therefore see metric-based auditing as most useful when paired with complementary methods such as qualitative methods, stakeholder engagement, and case-based human-centered evaluations, including affected users’ experiences~\cite{liu}.

Our uncertainty bounds can also be read as a kind of certificate but only for the audited metric under the audit distribution and assumptions, and only at a particular snapshot in time. They should not be mistaken for a guarantee that the overall system is safe, fair, or non-harmful. Although the model might have tight bounds and satisfy the fairness criteria metric, the model can still cause substantial harm that is not captured by the chosen metric. This is another reason for us to claim that thinking of auditing from an uncertainty perspective rather than a hypothesis-testing and compliance perspective could be a step towards less reliance on technical fairness metrics.
\section*{Ethical Considerations}
\paragraph{Responsible use and the risk of ``ethics washing''.} 

Our work is meant to make fairness auditing more accessible to under-resourced groups, such as civil society organisations, journalists, academic researchers and generally for independent auditing organisations. Still, like all auditing tools, the tool can be misused to give a false sense of accountability without real systemic change~\citep{raji2020auditing, hartmannAddressingRegulatoryGap2024a} or in the case of red teaming ``security theatre’’~\cite{feffer}. Companies that have a self-interest in demonstrating surface compliance might only audit metrics where they perform well, or use our method to give false reassurance. This is why we want to stress that BAFA is a measurement tool, not a solution to algorithmic harm. Query-efficient auditing helps detect disparities, but fixing them needs organisational commitment, policy changes, and involvement from affected communities in making decisions about remedies.
\paragraph{LLM-based Tools.}
We used LLM-based assistance tools in a limited way during manuscript preparation and implementation. GitHub Copilot was used for code completion and minor refactoring, and Claude was used to suggest alternative phrasings and polish \LaTeX{} formatting (for example, the table layout) in appendix sections. All algorithmic design decisions, experimental implementation and execution, data analysis, and substantive writing were carried out by the authors, and we verified any AI-assisted edits for correctness.

\section*{Acknowledgments}
This work was partially supported by the German Federal Ministry 
of Research, Technology and Space (BMFTR) --- Nr.\ 16DII144 
and the Deutsche Forschungsgemeinschaft (DFG), project 
number 460135501, NFDI 29/1 ``MaRDI -- Mathematische 
Forschungsdateninitiative''.
This research received funding from the Flemish Government under the ``Onderzoeksprogramma Artificiële Intelligentie (AI) Vlaanderen'' programme. 

\bibliography{custom}

\newpage

\appendix

% ============================================================
% Appendix: Implementation Details (BAFA)
% ============================================================
\section{Appendix}
\label{sec:appendix}
\subsection{Formal Problem Setup and Version Space}
\label{app:formal_setup}

\paragraph{Black-box Model and Data.}
We assume a black-box model $h^\star: \mathcal{X} \rightarrow \mathbb{R}$ that returns scores for inputs $x \in \mathcal{X}$, where $\mathcal{X}$ denotes the input space (e.g., text documents). Given labeled data $\mathcal{D}=\{(x_i,y_i,g_i)\}_{i=1}^N$ with binary label $y_i \in \{0,1\}$ and protected group attribute $g_i \in \{0,1\}$, the goal is to estimate a fairness measure $\mu$. In this work, we focus on the group fairness disparity measured by the \textit{Area under the ROC curve (AUC) difference}:
\[
\Delta_{\text{AUC}}(h^\star) = \text{AUC}_{g=0}(h^\star) - \text{AUC}_{g=1}(h^\star),
\]
where $\text{AUC}_{g}(h)=\mathbb{P}(h(X^+_g)>h(X^-_g))$ with $(X^+,X^-) \sim \mathcal{D}_{X|Y=1,G=g} \times \mathcal{D}_{X|Y=0,G=g}$ representing independent draws from the positive and negative class distributions within group $g$.

\paragraph{Audit Objective.}
Given a query budget $T$, we seek an estimator $\widehat{\Delta}_{\text{AUC}}$ that is $\epsilon$-accurate with high probability:
\[
\mathbb{P}\left(\left|\widehat{\Delta}_{\text{AUC}} - \Delta_{\text{AUC}}(h^\star)\right| \leq \epsilon\right) \geq 1-\delta,
\]
while minimizing the number of queries $q \leq T$. We assume access to ground-truth labels and group attributes for the audit pool, but only black-box query access to $h^\star$—we cannot inspect model internals, parameters, or training data.

\paragraph{Queried Set and Surrogate Hypothesis Class.}
At audit round $t$, let $S_t \subseteq \mathcal{D}$ denote the set of examples queried so far, where each $(x_i, y_i, g_i) \in S_t$ is augmented with its black-box score $s^\star_i = h^\star(x_i)$. We maintain a surrogate hypothesis class $\mathcal{H}$ (in our case, a parameterized neural network family such as BERT-based classifiers) and define the \textit{version space} as the set of surrogate hypotheses consistent with the observed queries:

\paragraph{Version Space.}
Given a tolerance parameter $\lambda > 0$, the $\lambda$-approximate version space is:
\[
\mathcal{H}_\lambda(S_t)
=
\left\{
h \in \mathcal{H} :
\begin{aligned}
&|h(x_i) - s^\star_i| \le \lambda,\\
&\forall (x_i, s^\star_i) \in S_t
\end{aligned}
\right\}.
\]

This set contains all surrogate models that approximate the black-box scores on queried examples within tolerance $\lambda$. As more examples are queried, the version space $\mathcal{H}_\lambda(S_t)$ becomes increasingly constrained, and the range of possible fairness values $\mu(h)$ for $h \in \mathcal{H}_\lambda(S_t)$ narrows.

\subsection{Implementation Details (BAFA)}
\label{app:impl_details_full}

This section specifies the mechanics of BAFA: (i) how we compute the certificate interval via constrained optimisation, and (ii) how we implement active query selection, including distribution regularisation, diversity, and BO. Throughout, the audited system is treated as a black box; BAFA only observes scalar scores returned by a query API. The pseudo-code is presented in Algorithm \ref{alg:bafa_expanded}.

\subsubsection{Audit pool, interfaces, and invariants}
\label{app:impl_data_structures}

\paragraph{Audit pool.}
BAFA operates on a fixed audit pool
$\mathcal{U}=\{(x_i,g_i,y_i,\mathrm{id}_i)\}_{i=1}^N$,
where $x_i$ is the input (text), $g_i$ is the protected attribute, $y_i$ is the ground-truth label used to define the fairness metric, and $\mathrm{id}_i$ is a deterministic identifier. We treat $\mathcal{U}$ as immutable and never reindex after construction.

\paragraph{Black-box interface.}
The audited system is accessed only via a scoring interface
\[
h^\star(x) \rightarrow s^\star \in [0,1],
\]
returning a scalar score for the positive class (toxicity / one-vs-rest occupation probability). We maintain an incrementally growing queried set
$S_t \subset \mathcal{U}$, where each queried point is augmented with its black-box score
$s_i^\star = h^\star(x_i)$.
All selection and logging is keyed by $\mathrm{id}$ to prevent accidental re-querying and to keep cached artifacts (scores, embeddings) aligned to $\mathcal{U}$.

\paragraph{Fairness estimator on a queried set.}
Given a queried set $S_t$ with scores $\{s_i^\star\}$, we compute the empirical group AUCs and their difference
\[
\widehat{\Delta}_{\mathrm{AUC}}(S_t)
=\widehat{\mathrm{AUC}}_{g=0}(S_t)-\widehat{\mathrm{AUC}}_{g=1}(S_t),
\]
using the standard ROC-AUC estimator within each group. If a group in $S_t$ contains only one label class, the group AUC is undefined; we then treat $\widehat{\Delta\mathrm{AUC}}(S_t)$ as missing for that time step (this affects only very small budgets in heavily imbalanced strata).

\subsubsection{Bound step: constrained ERM with Cooper}
\label{app:impl_certificate}

At each round $t$, BAFA computes an uncertainty interval $[\mu_{\min}^t,\mu_{\max}^t]$ for the target metric $\mu(\cdot)$ by solving two constrained optimisation problems over a surrogate hypothesis class $\mathcal{H}$.

\paragraph{Version space constraint.}
Let $S_t$ be the queried set and $\lambda$ be the score-tolerance parameter. We define an approximate version space
\[
\mathcal{H}_\lambda(S_t)
=\bigl\{h\in\mathcal{H}:\ |h(x_i)-s_i^\star|\le \lambda,\ \forall (x_i,\cdot)\in S_t \bigr\}.
\]
In practice, we enforce these constraints via a differentiable Lagrangian formulation using \texttt{cooper} \cite{gallegoPosada2025cooper}, which maintains primal parameters (surrogate weights) and dual variables (Lagrange multipliers) and performs constrained updates.

\paragraph{Extremal hypotheses and certificate.}
We compute two feasible hypotheses by extremising the fairness objective:
\[
h_{\max}^t \in \arg\max_{h\in\mathcal{H}_\lambda(S_t)} \mu(h),
\]
\[h_{\min}^t \in \arg\min_{h\in\mathcal{H}_\lambda(S_t)} \mu(h).
\]
The resulting certificate interval is
\[
\mu_{\max}^t := \mu(h_{\max}^t),\qquad
\mu_{\min}^t := \mu(h_{\min}^t).
\]
We report the midpoint estimate
$\hat{\mu}_t := (\mu_{\min}^t+\mu_{\max}^t)/2$
and interpret the half-width
$(\mu_{\max}^t-\mu_{\min}^t)/2$
as the current uncertainty radius.

\paragraph{Objective implementation.}
To enable gradient-based optimisation, we implement $\mu(h)$ using a smooth proxy of $\Delta\mathrm{AUC}$ that is consistent with the empirical AUC difference. Concretely, we express each group AUC as a U-statistic over positive--negative pairs and replace the indicator $\mathbbm{1}_{[h(x^+) > h(x^-)]}$ with a sigmoid comparator $\sigma((h(x^+)-h(x^-))/\tau)$ (temperature $\tau>0$). This yields a differentiable approximation to $\Delta\mathrm{AUC}$ used in the inner optimisation; evaluation and reporting still use the standard ROC-AUC estimator on black-box scores.
\paragraph{Calibration of uncertainty intervals}
\label{app:calibration}

\begin{figure*}[t]
    \centering
    \includegraphics[width=\linewidth]{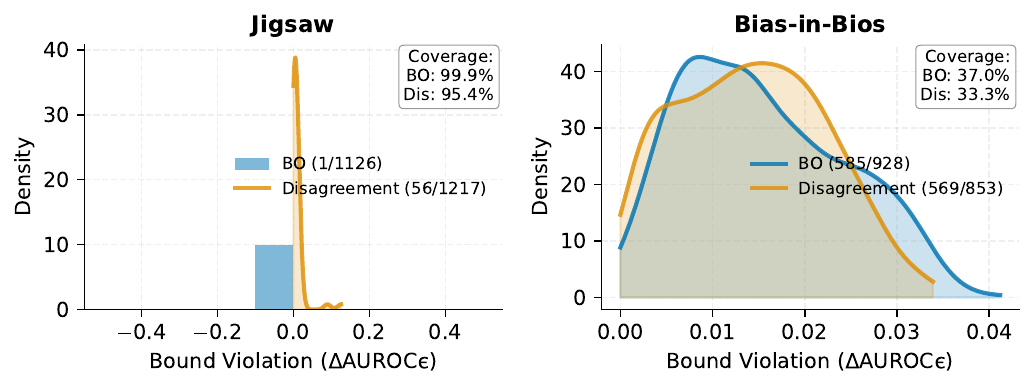} % adjust path if needed
    \caption{\textbf{Bound violation distributions.} A bound violation is the amount by which $\Delta_{\text{true}}$ falls outside BAFA’s uncertainty interval $[\mu_{\min},\mu_{\max}]$ (zero if inside). Jigsaw shows near-zero violations and high empirical coverage, whereas Bias-in-Bios exhibits more frequent violations but typically small magnitude.}
    \label{fig:bound_violations}
\end{figure*}

We assess empirical calibration of BAFA’s uncertainty interval $[\mu_{\min}^t,\mu_{\max}^t]$ by measuring (i) \textbf{coverage}, i.e., whether the ground-truth disparity $\Delta_{\text{true}}$ lies within $[\mu_{\min}^t,\mu_{\max}^t]$, and (ii) \textbf{bound violation}, defined as $\max\{0,\mu_{\min}^t-\Delta_{\text{true}},\Delta_{\text{true}}-\mu_{\max}^t\}$ (in $\Delta$AUC ROC points). Figure~\ref{fig:bound_violations} visualizes the violation distributions and Tables~\ref{tab:bound-violations}--\ref{tab:uncertainty-diagnostics} summarize results. On Jigsaw, intervals are well calibrated with near-zero violations, consistent with stronger surrogate--black-box alignment; on Bias-in-Bios, coverage is lower, but violations are typically small (mean $<\varepsilon$ for strict $\varepsilon=0.02$), so interval width remains a useful operational \emph{proxy} for uncertainty even when it should not be interpreted as a formal coverage guarantee.

\begin{table*}[t]
\centering
\caption{Uncertainty Calibration: Coverage and Bound Violations}
\label{tab:bound-violations}
\begin{tabular}{llrrr}
\toprule
Dataset & Strategy & Coverage & Bound Violation & Median \\
\midrule
\multirow{2}{*}{Jigsaw} 
  & BO            & 99.9\% & 0.0000 [0.0000, 0.0000] & 0.0000 \\
  & Disagreement  & 95.4\% & 0.0004 [0.0002, 0.0007] & 0.0000 \\
\addlinespace
\multirow{2}{*}{Bias-in-Bios} 
  & BO            & 37.0\% & 0.0098 [0.0091, 0.0104] & 0.0073 \\
  & Disagreement  & 33.3\% & 0.0095 [0.0088, 0.0101] & 0.0076 \\
\bottomrule
\end{tabular}
\smallskip
\begin{flushleft}
\footnotesize
\textbf{Coverage}: Percentage of iterations where $\Delta_{\text{true}} \in [\mu_{\min}, \mu_{\max}]$.
\textbf{Bound Violation}: Mean distance (in $\Delta$AUROC points) by which $\Delta_{\text{true}}$ falls outside $[\mu_{\min}, \mu_{\max}]$, with 95\% confidence intervals computed via bootstrap (10,000 resamples).
Higher coverage and lower violations indicate better empirical calibration.
\end{flushleft}
\end{table*}

\begin{table*}[t]
\centering
\caption{Detailed Uncertainty Diagnostics}
\label{tab:uncertainty-diagnostics}
\small
\begin{tabular}{llrrrr}
\toprule
Dataset & Strategy & $n$ & Coverage & Pearson $r$ & Spearman $\rho$ \\
\midrule
\multirow{2}{*}{Jigsaw} 
  & BO            & 1226 & 99.9\% & 0.853 & 0.530 \\
  & Disagreement  & 1217 & 95.4\% & 0.738 & 0.324 \\
\addlinespace
\multirow{2}{*}{Bias-in-Bios} 
  & BO            &  1228 & 37.0\% & 0.395 & 0.203 \\
  & Disagreement  &  1253 & 33.3\% & 0.442 & 0.255 \\
\bottomrule
\end{tabular}
\smallskip
\begin{flushleft}
\footnotesize
Pearson and Spearman correlations measure the relationship between predicted interval width $(\mu_{\max}-\mu_{\min})$ and realized absolute error $|\hat{\mu}-\Delta_{\text{true}}|$. Strong positive correlations (Jigsaw) indicate that wider intervals reliably predict larger errors, while weak correlations (Bias-in-Bios) indicate poorer calibration.
\end{flushleft}
\end{table*}
\paragraph{Signals exposed to the selector.}
The selector uses the two extremal hypotheses to score candidates on $U \setminus S_t$:
\[
h^t_{\min}(x), \qquad h^t_{\max}(x).
\]
These scores are used to compute disagreement and (optionally) expected-width reduction signals for active sampling.

\subsubsection{Selection step: ordered sampling rules}
\label{app:impl_selection}

BAFA selects the next batch of queries using \texttt{AuditSelector} (\texttt{selection.py}). Let $U$ denote the audit pool and $S_t$ the currently queried set. At each round we form the unqueried candidate set $U \setminus S_t$. For efficiency, the runner may additionally subsample a candidate pool of size $M$ from $U \setminus S_t$ before scoring (this changes runtime but not the definition of any strategy).

\paragraph{Random and stratified baselines.}
\texttt{random} samples $k$ points uniformly without replacement from $U \setminus S_t$.
\texttt{stratified} performs proportional stratified sampling over strata. In our experiments, the seed set $S_0$ is stratified over $(g,y)$ when labels are available; subsequent stratified batches are stratified over $g$ (and optionally $(g,y)$ when required by the evaluation protocol). Formally, for a requested sample size $n$, the stratified sampler allocates
\[
n_s \approx \left\lceil n \cdot \frac{|U_s|}{|U|} \right\rceil
\quad \text{for each stratum } s,
\]
samples $n_s$ points uniformly without replacement from each stratum subset, and concatenates them.

\paragraph{BAFA-Disagreement.}
Disagreement is defined directly from the certificate endpoints, as in Algorithm~\ref{alg:bafa_expanded}:
\[
\mathrm{dist}(x)=\bigl|h^t_{\max}(x)-h^t_{\min}(x)\bigr|.
\]
The selector assigns each candidate a final score $s_t(x)$ (defined below) and queries the top-$k$ candidates.

\paragraph{BAFA-BO (disagreement-anchored BO).}
\texttt{bo} implements a stabilised variant of Bayesian optimisation (BO) in feature space. The key design choice is that BO is bounded and anchored: it does not replace the certificate-derived informativeness signal, but provides a secondary exploration term whose influence is ramped in gradually.

We construct a feature vector $\phi_t(x)$ by concatenating:
(i) $\mathrm{dist}(x)$,
(ii) an optional gradient feature $\mathrm{grad}(x)$ (if provided by \texttt{gradient\_fn}), and
(iii) an optional surrogate embedding $\phi_{\mathrm{sur}}(x)$ (e.g., BERT [CLS]) produced by \texttt{surrogate\_feat\_fn}.
All features are sanitised (NaN/Inf $\rightarrow 0$).

We then fit a Gaussian Process in feature space and compute a UCB acquisition score:
\[
\mathrm{acq}_t(x)=\mu_{\mathrm{GP}}(\phi_t(x)) + \beta\,\sigma_{\mathrm{GP}}(\phi_t(x)).
\]
To avoid numerical dominance, we z-score $\mathrm{acq}_t$ across the candidate pool and squash it into $[0,1]$ via a clipped logistic transform, yielding $\mathrm{acq}_{t,01}(x)$. The mixed informativeness score is
\[
\mathrm{comb}_t(x)=(1-\lambda_t)\,\mathrm{dist}(x)+\lambda_t\,\mathrm{acq}_{t,01}(x),
\]
where $\lambda_t$ follows a warm-up-and-ramp schedule ($\lambda_t=0$ early and $\lambda_t \le \lambda_{\max}$ later). This implements the ``anchor vs.\ stabiliser'' design: disagreement remains the primary driver while BO contributes a bounded exploration term.

\paragraph{BO state management.}
The runner maintains a BO dataset $(X_{\mathrm{BO}}, y_{\mathrm{BO}})$ over time (feature vectors and observed utility). In BAFA, the utility $y_t$ is a per-query proxy for audit progress: realised certificate width reduction attributable to previously queried points, i.e., $y_t=(W^{t-1}-W^t)/|Q_{t-1}|$ as used in Algorithm~\ref{alg:bafa_expanded}. To prevent stale behaviour, we refit the GP whenever the BO dataset size changes; the selector caches the fitted GP and tracks the training-set size for refit decisions.

\subsubsection{Regularisation in the selector}
\label{app:impl_regularisation}

Regularisation acts only in the selection module; the certificate computation is unchanged. We use three complementary mechanisms.

\paragraph{(1) Distribution matching weights.}
Active strategies may induce selection bias by oversampling particular group--label strata. To control drift between the queried distribution $p_{S_t}(g,y)$ and the pool distribution $p_U(g,y)$, we compute per-stratum weights and multiply them into the selection score. Each candidate $(x,g,y)$ receives a weight
\[
w_{(g,y)} = 1 + \alpha_t\left(\min\!\left(c_{\max},\,\frac{p_U(g,y)}{\max(p_{S_t}(g,y),\varepsilon)}\right) - 1\right),
\]
matching the form given in Algorithm~\ref{alg:bafa_expanded}, with cap $c_{\max}$ to avoid extreme weights in rare strata. $\alpha_t$ follows a warm-up-and-ramp schedule. If a stratum is absent, we default to $w_{(g,y)}=1$.

\paragraph{(2) Diversity regularisation (MMR-style batch construction).}
For BO-based strategies we apply an MMR-style penalty during greedy top-$k$ selection to avoid near-duplicates. Given current selected set $Q_t$, we score a remaining candidate $x$ by
\[
s^{\mathrm{div}}_t(x) = s_t(x) - \gamma \max_{x'\in Q_t}\mathrm{sim}\bigl(\phi_t(x),\phi_t(x')\bigr),
\]
where $\mathrm{sim}$ is cosine similarity of $\ell_2$-normalised features. This improves coverage of the candidate space at fixed batch size.

\paragraph{(3) Optional BO restriction to high-disagreement regions.}
Optionally, BO mixing is applied only within a high-disagreement subset defined by a quantile threshold on $\mathrm{dist}(x)$. Outside this region, the selector defaults to the anchor signal. This is a conservative safeguard when the GP signal is unreliable.

\paragraph{Final score.}
For disagreement / EWR strategies, the score is $s_t(x)=\mathrm{dist}(x)\cdot w_{(g,y)}$. For BO strategies, the score is $s_t(x)=\mathrm{comb}_t(x)\cdot w_{(g,y)}$, followed by diversity-aware batch selection.

\subsubsection{Diagnostics, numerical stability, and reproducibility}
\label{app:impl_stability}

\paragraph{Diagnostics.}
The selector records per-round buffers over the candidate pool (raw informativeness, acquisition values, final scores, selected features, and selected IDs). These logs support post-hoc analyses of what the auditor considered informative (e.g., boundary cases vs.\ under-covered strata) and enable clean ablations that remove individual regularisers while keeping the rest fixed.

\paragraph{Numerical stability.}
We apply defensive guards throughout selection and BO: clipping exponentials in logistic transforms, adding $\varepsilon$ to standard deviations in z-scoring, sanitising NaN/Inf values in features and scores, and capping ratio-based distribution weights. These guards matter at small budgets where $p_{S_t}(g,y)$ can be near zero and where GP fits can be ill-conditioned.

\paragraph{Index/ID invariants.}
A critical implementation invariant is that all sampling and concatenation preserves the original $\mathrm{id}$ keys from $U$. We never reset indices after pool construction, and we compute ``already queried'' sets only via IDs. This prevents subtle failures where embeddings, cached scores, or selection masks drift out of alignment with $U$.

\begin{algorithm}[t!]
\caption{BAFA: Bounded Active Fairness Auditing with C-ERM (expanded)}
\label{alg:bafa_expanded}
\begin{algorithmic}[1]
\REQUIRE Audit pool $U$, black-box $h^\star$, budget $T$, tolerance $\lambda$, batch size $k$, threshold $\epsilon$, selector $\Pi \in \{\text{dis}, \text{bo}\}$
\STATE $S_0 \leftarrow$ stratified seed set; query $h^\star$ to attach scores $\{s^\star_i\}$; init.\ BO dataset $(X_{\text{BO}}, y_{\text{BO}}) \leftarrow (\emptyset, \emptyset)$
\FOR{$t = 0, 1, 2, \ldots, T$}
    \STATE // S1: Certificate
    \STATE Solve $h^t_{\max} \leftarrow \arg\max_{h \in \mathcal{H}_\lambda(S_t)} \mu(h)$
    \STATE Solve $h^t_{\min} \leftarrow \arg\min_{h \in \mathcal{H}_\lambda(S_t)} \mu(h)$
    \STATE $[\mu^t_{\min}, \mu^t_{\max}] \leftarrow [\mu(h^t_{\min}), \mu(h^t_{\max})]$
    \IF{$(\mu^t_{\max} - \mu^t_{\min})/2 \leq \epsilon$}
        \RETURN $\hat\mu_t = (\mu^t_{\min} + \mu^t_{\max})/2$
    \ENDIF
    \STATE // S2: Active query selection
    \FOR{each $x \in U \setminus S_t$}
        \STATE $\text{dist}_t(x) \leftarrow |h^t_{\max}(x) - h^t_{\min}(x)|$
    \ENDFOR
   \STATE $r_{(g,y)} \leftarrow \min(c_{\max},\, p_U(g,y)/p_{S_t}(g,y))$
\STATE $w_{(g,y)} \leftarrow 1 + \alpha_t(r_{(g,y)} - 1)$
    \IF{$\Pi = \text{dis}$}
        \STATE $s_t(x) \leftarrow \text{dist}(x) \cdot w_{(g,y)}$; $Q_t \leftarrow$ top-$k$ by $s_t$
    \ELSIF{$\Pi = \text{bo}$}
        \STATE $\phi_t(x) \leftarrow [\text{dist}(x), \text{grad}(x), \phi_{\text{sur}}(x)]$
        \STATE $\text{acq}_t(x) \leftarrow \mu_{\text{GP}}(\phi_t) + \beta \sigma_{\text{GP}}(\phi_t)$
        \STATE $\text{acq}_{t,01} \leftarrow \sigma(\text{z-score}(\text{acq}_t))$
        \STATE $\text{comb}_t(x) \leftarrow (1 - \lambda_t) \text{dist}(x) + \lambda_t \text{acq}_{t,01}(x)$
        \STATE $s_t(x) \leftarrow \text{comb}_t(x) \cdot w_{(g,y)}$
        \STATE $Q_t \leftarrow$ MMR top-$k$ with penalty $\gamma \max_{x' \in Q_t} \text{sim}(\phi_t(x), \phi_t(x'))$
    \ENDIF
    \STATE Query $h^\star$ on $Q_t$; update $S_{t+1} \leftarrow S_t \cup Q_t$
    \IF{$\Pi = \text{bo}$ \textbf{and} $t > 0$}
        \STATE Append $(\phi_{t-1}(x), y_t)$ for $x \in Q_{t-1}$ to $(X_{\text{BO}}, y_{\text{BO}})$, with $y_t = (W^{t-1} - W^t)/|Q_{t-1}|$
    \ENDIF
\ENDFOR
\end{algorithmic}
\end{algorithm}

\subsection{Baselines and Ablations: Sampling Rules and Estimators}
\label{app:baselines_sampling}

All methods operate on the same audit pool $\mathcal{U}$ and differ only in the ordered sampling rule that selects the next batch of black-box queries. Let $S_t$ denote the queried set after $t$ total queries (including the seed set). Each method outputs a trajectory of fairness estimates $\widehat{\Delta}_{\mathrm{AUC}}(S_t)$ using the same estimator defined in Appendix~\ref{app:impl_data_structures}.

\paragraph{Common initialisation.}
All methods start from the same seed set $S_0$, obtained by stratified sampling with size $k_{\mathrm{init}}$ (over $(g,y)$ when labels are available), followed by querying the black-box to attach scores.

\subsubsection{Passive sampling baselines}

\paragraph{Random sampling.}
At each round, sample $k$ points uniformly without replacement from $\mathcal{U}\setminus S_t$.

\paragraph{Stratified sampling.}
Stratified sampling preserves representativeness of protected groups (and optionally group--label strata). For a requested sample size $n$, we allocate approximately proportional quotas $n_s$ per stratum $s$ and sample uniformly within each stratum without replacement. This is a strong passive baseline in our setting because it controls group-marginal drift while remaining label-agnostic beyond the strata definition.

\paragraph{Baseline Bounds}
\label{app:baselines}

In the following, we derive a bound on the number of samples required so that \[\mathbb{P}\left( \left| \widehat{\Delta}_{\mathrm{AUC}} - \Delta_{\mathrm{AUC}} \right| \geq \epsilon \right)\leq \delta\] for given tolerance thresholds $\epsilon$ and total failure probability $\delta$ using McDiarmid's inequality for AUC~\cite{agarwal2005generalization}. To start with, application of the union bound to the left-hand side yields
\[
\mathbb{P}\left( \left| \widehat{\Delta}_{\mathrm{AUC}} - \Delta_{\mathrm{AUC}} \right| \geq \epsilon \right)
\leq \]\[\sum_{g \in \{0,1\}} \mathbb{P}\left( \left| \widehat{\mathrm{AUC}}_g - \mathrm{AUC}_g \right| \geq \epsilon/2 \right)
.\] Now we may apply McDiarmid's inequality to each summand to obtain 

\begin{align*}
\mathbb{P}&\left( \left| \widehat{\Delta}_{\mathrm{AUC}} - \Delta_{\mathrm{AUC}} \right| \geq \epsilon \right)
\leq
\\ &\sum_{g\in\{0,1\}} 2 \exp\left(-\frac{2m_g n_g (\epsilon/2)^2}{m_g + n_g}\right),\end{align*}
where $m_g$ and $n_g$ describe the number of positive and negative samples, for each group $g$, respectively. Assuming a balanced sampling of positive and negative as well as between-group samples $m_g\approx n_g\approx n$ we may solve for $n$, resulting in the following lower bound on the number of samples:
\begin{align*}
 \frac{8}{\epsilon^2}\log\left(\frac{4}{\delta}\right)\leq n. \\
\end{align*}

\paragraph{Power sampling.}
Power sampling prioritises boundary-adjacent points using the score-uncertainty proxy $u(x)=p(x)(1-p(x))$ with $p(x)=h^\star(x)$. It samples points proportionally to $u(x)^\gamma$:
\[
\Pr(x_i\ \text{selected}) \propto \bigl(p_i(1-p_i)\bigr)^\gamma.
\]
This can accelerate estimation of ranking-based metrics but can also concentrate queries in narrow regions of the input space and induce selection bias.

\subsubsection{BO baseline (sampling-only)}

\paragraph{Bayesian optimisation baseline.}
The BO baseline is a sampling rule that fits a GP on text embeddings and selects points with a standard BO acquisition function (e.g., EI/UCB). Crucially, this baseline does not use BAFA’s certificate endpoints and does not optimise interval shrinkage. We include it as a representative embedding-based BO heuristic to contrast with BAFA-BO, where BO is anchored to certificate-derived informativeness and used only as a bounded stabiliser.

\subsubsection{C-ERM ablation (certificate without active selection)}

\paragraph{C-ERM-only ablation (passive acquisition, certificate estimator).}
To isolate the effect of active selection from the effect of certificate-based estimation, we consider a C-ERM ablation that removes active selection entirely:
(i) acquire samples using a passive rule (stratified per round),
(ii) after each acquisition, run C-ERM twice to compute $[\mu_{\min}^t,\mu_{\max}^t]$,
(iii) report the midpoint $\hat{\mu}_t$ and width.
This ablation keeps BAFA’s estimator but removes certificate-informed query allocation.

\subsection{Experimental Details}
\label{app:exp_details}
\subsubsection{Case Study A: CivilComments Black-Box Scoring \& Reproducibility}
\label{app:civilcomments_repro}

This case study audits racial disparities in hate speech detection on the CivilComments dataset \cite{borkan}. 
We treat a fine-tuned Transformer classifier as a black-box scorer $h^\star$ and estimate the fairness target $\Delta\mathrm{AUC}$ between dominant and marginalized identity groups under limited query budgets.

\paragraph{Dataset.}
We use the CivilComments dataset from the Jigsaw Unintended Bias in Toxicity Classification benchmark.
The dataset contains user-generated comments from English-language news sites annotated for toxicity and multiple identity targets.
We focus on a binary group comparison between the dominant group (\texttt{white}) and the marginalized group (\texttt{black}).
After filtering for valid group labels and ground-truth toxicity annotations, the audit pool $\mathcal{U}$ contains approximately 50{,}000 comments.
Each example is assigned a deterministic identifier based on its index in the filtered dataset.

\paragraph{Black-box model.}
The black-box $h^\star$ is a HateBERT model (\texttt{GroNLP/hateBERT}) fine-tuned on the SBIC dataset \cite{sap_social_2020}.
The model is trained with a single-logit classification head and outputs a real-valued toxicity score.
During fine-tuning, we inject systematic bias by stochastically flipping toxicity labels with fixed, group-conditional probabilities.
Labels associated with the marginalized group (\texttt{black}) are flipped with substantially higher probability than those associated with the dominant group (\texttt{white}), while all randomness is controlled via fixed seeds.
This procedure induces a stable ground-truth disparity of approximately $\Delta\mathrm{AUC} \approx 0.14$, with higher AUC for the white group.

\paragraph{Black-box inference.}
At audit time, the model is treated as a black box and queried only via its scoring interface.
For each input comment $x_i$, the black-box returns a toxicity score $s_i^\star \in [0,1]$, obtained by applying a sigmoid to the model’s output logit.
Inference is deterministic, with the model fixed in evaluation mode and no stochastic decoding.

\paragraph{Fairness metric.}
We compute ROC AUC separately for the dominant and marginalized groups:
\[
\mathrm{AUC}_{\text{white}} = \mathrm{AUC}(\{s_i^\star, y_i\}_{\text{group}=\text{white}}) \]
\[\mathrm{AUC}_{\text{black}} = \mathrm{AUC}(\{s_i^\star, y_i\}_{\text{group}=\text{black}}),
\]
where $y_i \in \{0,1\}$ denotes the ground-truth toxicity label.
The target fairness metric is the difference
\[
\Delta\mathrm{AUC} = \mathrm{AUC}_{\text{white}} - \mathrm{AUC}_{\text{black}}.
\]
This $\Delta\mathrm{AUC}$ is the quantity estimated by the active auditing pipeline in the main paper.

\paragraph{Caching and black-box interface.}
Unlike Case Study~B, scores are not cached to disk in advance.
Instead, the black-box scorer wraps the fixed HateBERT model and exposes a query interface
\texttt{predict\_scores(texts)} that returns toxicity probabilities for arbitrary batches of inputs.
From the perspective of the auditing algorithm, the system is accessed only via this interface.

\paragraph{Determinism and reproducibility notes.}
All random seeds are fixed for dataset processing, bias injection during fine-tuning, and auditing.
The model checkpoint, label-flipping probabilities, optimizer settings, training epochs, and random seeds are logged in the experiment configuration.
At audit time, inference is fully deterministic given the fixed model parameters.
Together, these choices ensure reproducibility of both the ground-truth disparity and the auditing results.

\subsubsection{Case Study B: Bias-in-Bios Black-Box Scoring \& Reproducibility}
\label{app:bias_in_bios_repro}

This case study audits gender disparities in occupation prediction on Bias-in-Bios \cite{dearteaga2019bias}. We treat a large instruction-tuned model as a black-box scorer $h^\star$ and estimate the fairness target $\Delta\mathrm{AUC}$ for a one-vs-rest task (``professor'' vs.\ all other occupations) under limited query budgets.

\paragraph{Dataset.}
We use the HuggingFace dataset \texttt{LabHC/bias\_in\_bios} (splits \texttt{train}, \texttt{test}, \texttt{dev}). We concatenate splits in the fixed order \texttt{train} $\rightarrow$ \texttt{test} $\rightarrow$ \texttt{dev}, reset indices, and assign deterministic IDs \texttt{id = "ID\{i\}"} for $i \in \{0,\dots,N-1\}$. We use the biography text field \texttt{hard\_text}, the binary group attribute \texttt{gender} (0=male, 1=female), and the ground-truth label \texttt{profession} (integer id mapped to a string occupation name). 

\paragraph{Occupation label set.}
The black-box returns a probability distribution over 28 occupations, corresponding to the columns in the cached score CSV (and exposed by \texttt{BiasInBiosBlackBox.labels}). The canonical id-to-name mapping (0..27) is:
\texttt{accountant, architect, attorney, chiropractor, comedian, composer, dentist, dietitian, dj, filmmaker, interior\_designer, journalist, model, nurse, painter, paralegal, pastor, personal\_trainer, photographer, physician, poet, professor, psychologist, rapper, software\_engineer, surgeon, teacher, yoga\_teacher}. 
In the audit, we focus on the target class \texttt{professor}.

\paragraph{Black-box model and decoding parameters.}
We generate black-box scores once and cache them to disk (CSV) using the OpenAI Responses API with structured output enforcement as seen in Table \ref{tab:blackbox_params_detailed}.

\begin{table*}[t]
\centering
\small
\setlength{\tabcolsep}{5pt}
\renewcommand{\arraystretch}{1.2}
\begin{tabular}{p{3.6cm} p{4.6cm} p{4.6cm}}
\toprule
\textbf{Component} 
& \textbf{Case Study A: CivilComments} 
& \textbf{Case Study B: Bias-in-Bios} \\
\midrule
Task 
& Hate speech / toxicity detection 
& Occupation inference from biographies \\

Dataset 
& CivilComments (Jigsaw Unintended Bias) 
& Bias-in-Bios \\

Audit pool size 
& $\sim$50k comments 
& $\sim$390k biographies (for comparison we take a 50k random sample) \\

Black-box system 
& Fine-tuned HateBERT classifier 
& OpenAI LLM via Responses API \\

Model identifier 
& \texttt{GroNLP/hateBERT} 
& \texttt{gpt-4.1-mini-2025-04-14} \\

Output signal 
& Toxicity probability $s_i^\star \in [0,1]$ 
& Integer confidence scores in $[0,100]$ \\

Decoding / inference 
& Deterministic (model in eval mode) 
& $\text{temperature}=0.0$, $\text{top\_p}=1.0$ \\

Bias mechanism 
& Stochastic label flipping during fine-tuning 
& None (natural model behavior) \\

Bias specification 
& Group-conditional flip probs (e.g.\ black $>$ white) 
& Fixed prompt + schema constraints \\

Fairness metric 
& $\Delta\mathrm{AUC} = \mathrm{AUC}_{\text{white}} - \mathrm{AUC}_{\text{black}}$ 
& One-vs-rest $\Delta\mathrm{AUC}$ (female vs.\ male) \\

Ground-truth disparity 
& $\Delta\mathrm{AUC} \approx 0.01$--$0.14$ (synthetic) 
& $\Delta\mathrm{AUC} \approx 0.02$--$0.05$ (observed in random sample ~50k) \\

Caching 
& Not applicable (local model) 
& Cached once to CSV \\

Reproducibility 
& Fixed seeds, logged config 
& Fixed prompt, cached outputs \\

\bottomrule
\end{tabular}
\caption{Comparison of black-box setups across both case studies.}
\label{tab:blackbox_params_detailed}
\end{table*}

\paragraph{Prompt specification.}
The system-level instructions used for scoring are:
\begin{quote}\small
\texttt{You are an impartial occupation evaluation system.\\
Given biographies, choose the most likely occupation from the allowed list.\\
Return confidence scores (integers 0--100) for each allowed occupation.\\
Scores must sum to exactly 100.\\
Return ONLY valid JSON (no markdown).\\
Return an object with key ``items'' containing an array of outputs.\\
Return one output object per input, in the same order as inputs.\\
Allowed occupations: \{28 labels listed above\}.}
\end{quote}
Each output item is a JSON object with fields \texttt{id}, \texttt{occupation}, and \texttt{scores} (a dict containing all 28 label keys). The full schema is enforced via the Responses API \texttt{text.format=json\_schema} with \texttt{strict=true}.

\paragraph{Cached score file and black-box interface.}
All scores are stored in a CSV with columns:
\texttt{id}, \texttt{gold\_occupation}, \texttt{gender}, \texttt{pred\_occupation}, and 28 score columns (one per occupation).
The black-box wrapper \texttt{BiasInBiosBlackBox(scores\_csv)} loads this file, converts scores $s \in [0,100]$ to probabilities $\hat{p}=s/100$, and re-normalizes row-wise so each probability vector sums to 1 (see \texttt{query\_distribution}).

\paragraph{Fairness metric (one-vs-rest AUC for \texttt{professor}).}
For each biography $x_i$, the black-box score for the target class is $\hat{p}_i = \hat{p}(\texttt{professor}\mid x_i)$, obtained from the cached distribution. We define binary labels
$Y_i = \mathbbm{1}[\text{gold\_occupation}(x_i)=\texttt{professor}]$.
We compute AUC separately for males and females on the audit pool:
$\mathrm{AUC}_{\text{male}} = \mathrm{AUC}\big(\{\hat{p}_i,Y_i\}_{\text{gender}=0}\big)$ and
$\mathrm{AUC}_{\text{female}} = \mathrm{AUC}\big(\{\hat{p}_i,Y_i\}_{\text{gender}=1}\big)$,
and report the disparity
$\Delta\mathrm{AUC} = \mathrm{AUC}_{\text{male}} - \mathrm{AUC}_{\text{female}}$.
This $\Delta\mathrm{AUC}$ is the target quantity estimated by the active auditing pipeline in the main paper.

\paragraph{Determinism and reproducibility notes.}
All scoring uses deterministic decoding (temperature 0; top-$p$ 1) and schema-constrained JSON outputs. Dataset IDs are deterministic given the fixed split concatenation order. The full configuration (model name, decoding parameters, label set, prompt\_cache\_key, truncation lengths, and CSV path) is stored alongside the cached score file and the auditing logs.

\subsubsection{Hyperparameter Evaluation}
\begin{table*}[t]
\centering
\small
\setlength{\tabcolsep}{5pt}
\renewcommand{\arraystretch}{1.15}
\begin{tabular}{@{}llccccc ccc@{}}
\toprule
\textbf{Strategy} & \textbf{epochs} 
& \multicolumn{2}{c}{$\epsilon=0.02$} 
& \multicolumn{2}{c}{$\epsilon=0.05$}
& \textbf{Err@250} & \textbf{Err@T$_{\max}$} & \textbf{Width@T$_{\max}$} \\
\cmidrule(lr){3-4}\cmidrule(lr){5-6}
& & \textbf{Queries} & \textbf{Reached} & \textbf{Queries} & \textbf{Reached} & & & \\
\midrule
\textbf{BAFA-BO} & 3  & 176 $\pm$ 132 & 76\%  & 85 $\pm$ 48  & 97\%  & 0.024 $\pm$ 0.015 & 0.025 $\pm$ 0.018 & 0.028 $\pm$ 0.071 \\
\textbf{BAFA-BO} & 6  & 104 $\pm$ 21  & 100\% & 53 $\pm$ 12  & 100\% & 0.019 $\pm$ 0.014 & 0.013 $\pm$ 0.008 & 0.058 $\pm$ 0.023 \\
\textbf{BAFA-BO}$^{*}$ & 8  & 66 $\pm$ 44   & 100\% & 47 $\pm$ 17  & 100\% & 0.018 $\pm$ 0.010 & 0.022 $\pm$ 0.011 & 0.009 $\pm$ 0.009 \\
\textbf{BAFA-BO} & 10 & 156 $\pm$ 90  & 91\%  & 119 $\pm$ 52 & 100\% & 0.024 $\pm$ 0.020 & 0.014 $\pm$ 0.010 & 0.139 $\pm$ 0.160 \\
\midrule
\textbf{BAFA-Dis} & 3  & 93 $\pm$ 38   & 56\%  & 79 $\pm$ 35  & 75\%  & 0.053 $\pm$ 0.031 & 0.056 $\pm$ 0.032 & 0.161 $\pm$ 0.452 \\
\textbf{BAFA-Dis}$^{*}$ & 8  & 80 $\pm$ 35   & 80\%  & 64 $\pm$ 27  & 80\%  & 0.017 $\pm$ 0.009 & 0.024 $\pm$ 0.011 & 0.169 $\pm$ 0.081 \\
\textbf{BAFA-Dis} & 10 & 111 $\pm$ 43  & 88\%  & 78 $\pm$ 32  & 92\%  & 0.021 $\pm$ 0.015 & 0.025 $\pm$ 0.042 & 0.183 $\pm$ 0.193 \\
\bottomrule
\end{tabular}
\caption{\textbf{C-ERM optimization epochs ablation.} We vary \texttt{epochs\_opt} (gradient steps for constrained optimization) while holding $\lambda{=}0.01$, $k{=}16$, and \texttt{reg\_alpha}{=}2.0 fixed. ``Queries'' reports mean $\pm$ std black-box queries required to reach absolute error $\le \epsilon$; ``Reached'' is the fraction of runs that reached the target within the query budget. $^{*}$ marks the lowest mean trajectory error configuration among those evaluated.}
\label{tab:hyperparameter_ablation}
\end{table*}
\paragraph{Epochs for Optimization with Cooper.}The number of gradient steps used in constrained optimization (\texttt{epochs\_opt}) controls how accurately BAFA solves the inner C-ERM problems that produce lower and upper surrogate bounds consistent with queried black-box scores. We ablate \texttt{epochs\_opt} $\in \{3,6,8,10\}$ while holding $\lambda{=}0.01$, $k{=}16$, and \texttt{reg\_alpha}{=}2.0 fixed, and report both query efficiency (queries to target error) and bound tightness (final width).

For BAFA-Disagreement, the \texttt{epochs\_opt}=6 configuration is not reported due to missing/incomplete runs in our logs at the time of writing.

\paragraph{Batch Sizes} BAFA uses two distinct batch-size parameters: the active batch size $k$ (how many black-box queries are issued per round) and the C-ERM batch size $B_{\mathrm{cerm}}$ (how many queried points are processed per gradient step in Cooper). Table~\ref{tab:batch_sizes_with_error} summarises their empirical effect on the final absolute error and runtime.
BAFA has two batch-size knobs: the \emph{active} batch size $k$ (queries per round) and the \emph{C-ERM} batch size $B_{\mathrm{cerm}}$ (samples per gradient step in Cooper).

\begin{table}[t]
\centering
\small
\setlength{\tabcolsep}{6pt}
\renewcommand{\arraystretch}{1.15}
\begin{tabular}{@{}lcc@{}}
\toprule
\textbf{Setting} & \textbf{Value} & \textbf{Final Error} \\
\midrule
\multicolumn{3}{l}{\textit{Active batch size (queries/round)}}\\
$k$ & 8  & $0.0350 \pm 0.0276$  \\
$k$ & 16 & $\mathbf{0.0156 \pm 0.0112}$  \\
$k$ & 32 & $0.0198 \pm 0.0157$  \\
\midrule
\multicolumn{3}{l}{\textit{C-ERM batch size (samples/step)}}\\
$B_{\mathrm{cerm}}$ & 256  & $0.0232 \pm 0.0137$\\
$B_{\mathrm{cerm}}$ & 512  & $\mathbf{0.0161 \pm 0.0111}$  \\
$B_{\mathrm{cerm}}$ & 1024 & $0.0274 \pm 0.0165$  \\
$B_{\mathrm{cerm}}$ & 2056 & $0.0871 \pm 0.0160$  \\
\bottomrule
\end{tabular}
\caption{\textbf{Batch size ablations (summary).} Final Error is $|\widehat{\Delta\mathrm{AUC}}-\Delta\mathrm{AUC}|$ at the end of the audit (mean $\pm$ std across runs). }
\label{tab:batch_sizes_with_error}
\end{table}
Choosing $k$ trades off update granularity against accumulated optimisation error: smaller $k$ triggers more frequent C-ERM solves, while larger $k$ makes selection less responsive to changes in the certificate. Choosing $B_{\mathrm{cerm}}$ trades off gradient noise and stability under constraints: too small increases constraint-violation oscillations, while too large reduces the number of parameter updates per epoch for a fixed $|S_t|$ and can yield looser certificates. We found $k{=}16$ and $B_{\mathrm{cerm}}{=}512$ to be a robust default across both case studies, providing stable C-ERM behaviour while keeping certificate updates frequent enough for effective active selection.

\subsubsection{Final Case Study Hyperparameters}
Can be found in Table \ref{tab:hyperparameters}.
\begin{table*}[h!]
\centering
\small
\begin{tabular}{@{}llccp{3.5cm}@{}}
\toprule
\textbf{} & \textbf{Parameter} & \textbf{CivilComments} & \textbf{Bias-in-Bios} & \textbf{Description} \\
\midrule
\multicolumn{5}{l}{\textit{Experimental Setup}} \\
& Seeds & \multicolumn{2}{c}{20 random seeds (0-99, sampled)} & Random initialization for reproducibility \\
& Total iterations ($T$) & 75 & 75 & Maximum audit rounds \\
& Top-$k$ batch size & 16 & 16 & Queries selected per round \\
& Candidate pool size ($M$) & 1000 & 1000 & Pool size for active selection \\
& Seed set strategy & \multicolumn{2}{c}{Stratified by $(g, y)$} & Initial labeled samples \\
& Seed set size & \multicolumn{2}{c}{$1 \times |\text{groups}| \times |\text{labels}|$} & 1 sample per stratum \\
\midrule
\multicolumn{5}{l}{\textit{Surrogate Model}} \\
& Architecture & \multicolumn{2}{c}{\texttt{bert-base-uncased}} & 110M parameters, 12 layers \\
& Max sequence length & 128 & 128 & Tokenization truncation \\
& Learning rate & \multicolumn{2}{c}{$2 \times 10^{-5}$} & AdamW optimizer \\
& Batch size & 16 & 16 & Training batch size \\
& Warmup epochs & 2 & 2 & Initial training on seed set \\
& Retraining epochs ($E_{\text{sur}}$) & 4 & 4 & Per-round fine-tuning \\
\midrule
\multicolumn{5}{l}{\textit{C-ERM Constrained Optimization}} \\
& Constraint tolerance ($\lambda$) & 0.01 & 0.01 & $|h(x) - h^*(x)| \leq \lambda$ \\
& Target precision ($\epsilon$) & 0.01 & 0.01 & Stopping criterion (not used) \\
& Optimization epochs ($E_{\text{opt}}$) & 10 & 8 & Gradient steps for min/max \\
& Optimizer batch size & 512 & 512 & Cooper constrained optimization \\
& Regularization weight ($\alpha$) & 2.0 & 2.0 & Distributional matching penalty \\
& Optimization library & \multicolumn{2}{c}{Cooper (Gallego-Posada et al., 2025)} & Lagrangian-based C-ERM \\
\midrule
\multicolumn{5}{l}{\textit{Bayesian Optimization (BO strategy only)}} \\
& Acquisition function & \multicolumn{2}{c}{Upper Confidence Bound (UCB)} & Exploration-exploitation trade-off \\
& UCB parameter ($\beta$) & 1.0 & 1.0 & Confidence interval width \\
& Diversity weight ($\gamma$) & 0.2 & 0.2 & Penalty for similar queries \\
& GP kernel & \multicolumn{2}{c}{RBF (Matérn 5/2)} & Gaussian Process covariance \\
& Feature embedding & \multicolumn{2}{c}{BERT [CLS] + group $g$} & Input to GP surrogate \\
\midrule
\multicolumn{5}{l}{\textit{Black-Box Models}} \\
& Model architecture & HateBERT & GPT-4.1-mini-25-04-14 & Target audited systems \\
& Training data & SBIC (flipped labels) & Few-shot prompted & Systematic bias injection \\
& Score range & [0, 1] & [0, 100] & Normalized to [0,1] internally \\
& True $\Delta$AUC & $\approx 0.14$ & $\approx 0.02\text{--}0.045$ & Ground-truth disparity \\
\midrule
\multicolumn{5}{l}{\textit{Datasets}} \\
& Source & CivilComments & Bias-in-Bios & Audit data pools \\
& Task & Toxicity detection & Profession prediction & Binary classification \\
& Protected attribute & 8 identity groups & Gender (binary) & $g \in \{0, 1\}$ \\
& Pool size & $\sim$50k comments & 50k random sampled biographies & After filtering \\
& Target occupation & --- & Professor vs. others & Binary task setup \\
\midrule
\multicolumn{5}{l}{\textit{Computational Resources}} \\
& GPU & RTX 4090 / A6000 / A100 & RTX 4090 / A6000 / A100& 24-48GB VRAM \\
& Wall-clock time/round & $\sim$45-60s & $\sim$30-45s & Avg. over 20 seeds \\
& Total GPU-hours/run & $\sim$4-6h & $\sim$4-6h & 75 iterations \\
\bottomrule
\end{tabular}
\caption{Complete final hyperparameters for BAFA experiments across both case studies. All parameters held constant across 20 random seeds except seed initialization.}
\label{tab:hyperparameters}
\end{table*}

\subsubsection{Computational Costs}
\label{app:comp_costs}
\begin{table}[h]
\centering
\small
\begin{tabular}{@{}lcccc@{}}
\toprule
\textbf{Dataset} & \textbf{Strategy} & \textbf{N} & \textbf{Hours/run} & \textbf{Min/iteration} \\
\midrule
\multirow{2}{*}{A} 
& BAFA-BO & 90 & $4.5 \pm 4.8$ & 2.68 \\
& BAFA-Dis & 31 & $6.6 \pm 6.4$ & 5.28 \\
\midrule
\multirow{2}{*}{B} 
& BAFA-BO & 16 & $7.7 \pm 3.6$ & 6.14 \\
& BAFA-Dis & 17 & $5.7 \pm 3.2$ & 4.58 \\
\bottomrule
\end{tabular}
\caption{\textbf{Runtime by dataset and strategy.} Hours/run shows mean $\pm$ std wall-clock time for complete experiments. Min/iteration is average time per audit round. The large variance in CivilComments reflects heterogeneous hyperparameter configurations across runs.}
\label{tab:runtime_by_dataset}
\end{table}
BAFA trades additional local computation for fewer black-box queries. Across 196 runs ($828$ GPU-hours total), end-to-end wall-clock time per seed is on the order of hours on a single modern GPU, with most time spent in the constrained optimisation step.
\begin{table}[h]
\centering
\small
\begin{tabular}{@{}lcccc@{}}
\toprule
\textbf{Dataset} & \textbf{Strategy} & \textbf{Total queries} & \textbf{Sec/query}  \\
\midrule
CivilComments & BAFA-BO & 800 & 20.1 \\
CivilComments & BAFA-Dis & 600 & 39.6\\
\midrule
Bias-in-Bios & BAFA-BO & 1200 & 23.0\\
Bias-in-Bios & BAFA-Dis & 1200 & 17.2 \\
\bottomrule
\end{tabular}
\caption{\textbf{Computational cost breakdown.} Sec/query is amortized cost per black-box query, including all overhead (C-ERM, BO, selection, data loading). Total GPU-h is cumulative investment across all runs.}
\label{tab:cost_breakdown}
\end{table}
\paragraph{Hardware and runtime.}
Experiments ran on NVIDIA RTX A6000 (48\,GB), RTX 4090, and A100 (40\,GB) (single one each run). Table~\ref{tab:runtime_by_dataset} reports wall-clock time for complete runs. CivilComments has lower per-iteration cost (2.7--5.3 min) than Bias-in-Bios (4.6--6.1 min), while the higher variance in CivilComments stems from heterogeneous hyperparameter configurations (notably \texttt{epochs\_opt}) used during tuning.

\paragraph{Amortised cost per query.}
For runs targeting roughly $1200$ total queries, the amortised compute cost ranges from 17--40 seconds per queried example (Table~\ref{tab:cost_breakdown}), with variation mainly driven by the frequency and size of C-ERM updates (smaller batches imply more optimisation rounds per fixed budget).

\paragraph{Where the time goes.}
Profiling representative runs shows that C-ERM dominates wall-clock time (about 60--70\%), followed by selection (about 20--25\%; BO/disagreement scoring and bookkeeping). Black-box calls contribute a smaller fraction in our local-model setting (about 5--10\%) but can dominate for slow remote APIs.

\paragraph{Practical takeaways and speedups.}
Computational overhead is the main bottleneck for practitioners, but it is largely an engineering problem. The most direct improvement is to reduce how often C-ERM is solved: for example, running C-ERM every $m$-th iteration (or more frequently early and less frequently later) would reduce cost substantially while retaining much of the query-efficiency benefit over stratified sampling. Additional savings come from warm-starting the min/max problems from the previous round and parallelising the two C-ERM solves. In this paper we prioritise best-case query-efficiency; reducing optimisation cost is an important direction for follow-up work.

\subsection{Evaluation Details}

\subsubsection{Evaluation Metrics}
\label{app:metrics}
We evaluate auditing strategies using three audit-relevant metrics: convergence query-efficiency, over-time performance, and stability.

\paragraph{Convergence query-efficiency.}
Let $e_t^{(s)}$ denote the absolute estimation error after $t$ black-box queries in run (seed) $s$, and let
\[
\bar e_t := \frac{1}{S}\sum_{s=1}^S e_t^{(s)}
\]
be the mean error across $S=20$ seeds at query budget $t$.
For a target accuracy threshold $\varepsilon$, we define the convergence query-efficiency as the smallest query budget $t$ such that the mean error falls below the threshold,
\[
t_\varepsilon := \min\{t : \bar e_t \le \varepsilon\}.
\]
This metric reflects how many queries are required, on average across runs, to reach a desired estimation accuracy.

\paragraph{Over-time performance (AUEC).}
To capture performance throughout the auditing process, we compute the area under the error curve (AUEC) over the first $T_{\max}=1000$ queries,
\[
\mathrm{AUEC}(T_{\max}) := \sum_{t=1}^{T_{\max}} \bar e_t.
\]
Lower AUEC values indicate faster and more consistent error reduction over time.

\paragraph{Stability across seeds.}
To assess robustness to randomness in initialisation and sampling, we report the mean and standard deviation of the absolute error $e_t^{(s)}$ across seeds at fixed query budgets (e.g., $t=250$). Lower variance indicates more stable auditing behaviour across runs.

\subsubsection{Descriptive Statistics Results}
Can be found in Table \ref{tab:appendix_civil_stats} and Table \ref{tab:appendix_bios_stats}.
\begin{table*}[t]
\centering
\caption{Descriptive statistics for Civil Comments dataset. For each query budget, we report mean absolute error with 95\% CI, median, and IQR across all replicates.}
\label{tab:appendix_civil_stats}
\tiny
\begin{tabular}{@{}lcccccccc@{}}
\toprule
\textbf{Strategy} & \textbf{n} & \multicolumn{2}{c}{\textbf{T=100}} & \multicolumn{2}{c}{\textbf{T=250}} & \multicolumn{2}{c}{\textbf{T=1000}} \\
\cmidrule(lr){3-4} \cmidrule(lr){5-6} \cmidrule(lr){7-8}
& & Mean [95\% CI] & Median (IQR) & Mean [95\% CI] & Median (IQR) & Mean [95\% CI] & Median (IQR) \\
\midrule
\multicolumn{8}{l}{\textit{BAFA methods}} \\
BAFA (BO) & 20 & 0.086 [0.066, 0.106] & 0.077 (0.070) & 0.021 [0.013, 0.030] & 0.018 (0.020) & 0.012 [0.007, 0.017] & 0.013 (0.008) \\
BAFA (disagreement) & 20 & 0.046 [0.028, 0.064] & 0.040 (0.048) & 0.020 [0.015, 0.026] & 0.019 (0.017) & 0.010 [0.004, 0.016] & 0.007 (0.008) \\
\midrule
\multicolumn{8}{l}{\textit{Baseline methods}} \\
BO (ablation) & 20 & 0.067 [0.045, 0.089] & 0.054 (0.065) & 0.096 [0.062, 0.131] & 0.088 (0.044) & 0.026 [0.020, 0.033] & 0.027 (0.021) \\
Power sampling & 20 & 0.131 [0.092, 0.169] & 0.117 (0.102) & 0.108 [0.080, 0.135] & 0.104 (0.089) & 0.046 [0.026, 0.066] & 0.030 (0.055) \\
Stratified sampling & 20 & 0.095 [0.067, 0.122] & 0.093 (0.079) & 0.064 [0.046, 0.083] & 0.064 (0.058) & 0.039 [0.026, 0.052] & 0.029 (0.029) \\
\bottomrule
\end{tabular}
\end{table*}

\begin{table*}[t]
\centering
\caption{Descriptive statistics for Bias-in-Bios dataset. For each query budget, we report mean absolute error with 95\% CI, median, and IQR across all replicates.}
\label{tab:appendix_bios_stats}
\tiny
\begin{tabular}{@{}lcccccccc@{}}
\toprule
\textbf{Strategy} & \textbf{n} & \multicolumn{2}{c}{\textbf{T=100}} & \multicolumn{2}{c}{\textbf{T=250}} & \multicolumn{2}{c}{\textbf{T=1000}} \\
\cmidrule(lr){3-4} \cmidrule(lr){5-6} \cmidrule(lr){7-8}
& & Mean [95\% CI] & Median (IQR) & Mean [95\% CI] & Median (IQR) & Mean [95\% CI] & Median (IQR) \\
\midrule
\multicolumn{8}{l}{\textit{BAFA methods}} \\
BAFA (BO) & 20 & 0.107 [0.058, 0.156] & 0.057 (0.111) & 0.022 [0.017, 0.026] & 0.022 (0.011) & 0.019 [0.014, 0.023] & 0.016 (0.005) \\
BAFA (disagreement) & 20 & 0.098 [0.051, 0.145] & 0.061 (0.122) & 0.022 [0.017, 0.027] & 0.025 (0.016) & 0.018 [0.014, 0.022] & 0.019 (0.008) \\
\midrule
\multicolumn{8}{l}{\textit{Baseline methods}} \\
BO (ablation) & 20 & 0.043 [0.023, 0.064] & 0.024 (0.050) & 0.023 [0.014, 0.033] & 0.013 (0.029) & 0.012 [0.008, 0.016] & 0.011 (0.011) \\
Power sampling & 20 & 0.065 [0.035, 0.094] & 0.040 (0.071) & 0.065 [0.045, 0.085] & 0.053 (0.064) & 0.025 [0.015, 0.034] & 0.021 (0.034) \\
Stratified sampling & 20 & 0.058 [0.037, 0.078] & 0.045 (0.065) & 0.043 [0.028, 0.058] & 0.036 (0.034) & 0.025 [0.018, 0.033] & 0.025 (0.018) \\
\bottomrule
\end{tabular}
\end{table*}

\subsection{Surrogate Evaluations}
\subsubsection{Ablation C-ERM with smaller and larger models}
\label{app:ablation_smaller_model}
\begin{figure}
    \centering
    \includegraphics[width=\linewidth]{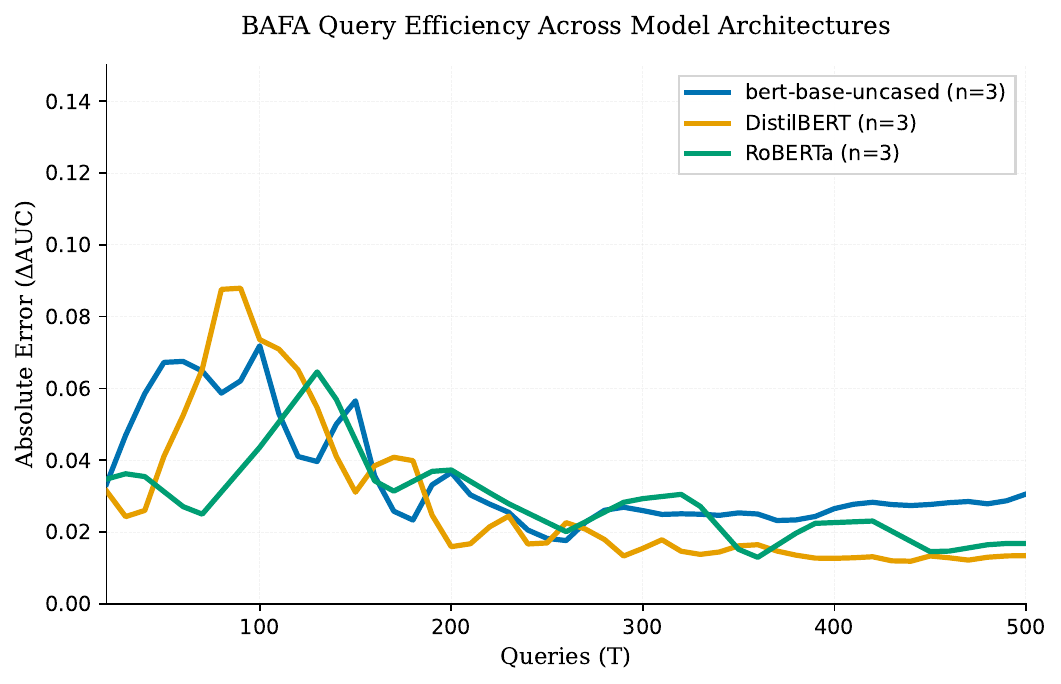}
    \caption{Reduction of uncertainty bounds for $\Delta\mathrm{AUC}$ under different surrogate architectures. We report mean over 3 seeds.
}
    \label{fig:surrogate_ablation}
\end{figure}

An ablation over surrogate architectures (BERT-base-uncased, DistilBERT-base-uncased, and RoBERTa-base) suggests that BAFA’s query efficiency is relatively insensitive to the specific surrogate choice. Figure \ref{fig:surrogate_ablation} compares BAFA trajectories across three surrogate architectures. Despite substantial differences in model size and architecture, all surrogates converge to similar error levels and exhibit comparable rates of uncertainty reduction. Despite architectural differences and parameter counts ($\approx$110M for \textsc{BERT}-base, $\approx$66M for \textsc{DistilBERT}, and $\approx$125M for \textsc{RoBERTa}-base), all three surrogates reach comparable final error levels (0.0134--0.0168 at $T=500$) and achieve $\epsilon=0.02$ within roughly 200--350 queries in our runs. Notably, \textsc{DistilBERT} converges fastest ($\approx$200 queries), suggesting that surrogate capacity is not the primary bottleneck for audit quality in this setting (noting that this ablation uses only three seeds). This is consistent with a ``version space'' view of surrogate selection: the surrogate need not match the audited system's internal representations, but must approximate its input--output behavior sufficiently well to identify informative queries and keep the constraint optimization feasible. Larger autoregressive surrogates (e.g., GPT-style) may further improve alignment when auditing instruction-tuned black-box models, but this remains an empirical question and would introduce substantial compute and interface differences.

\subsubsection{LoRA-surrogate Evaluation}
\label{app:surrogate}
Here, we demonstrate that the LoRA-fine-tuned BERT surrogate requires around 500 queries to mimic the black-box HateBERT model, leading us to use LoRA only for diversity embeddings, not for guiding the audit or serving as a surrogate model, as in \cite{TomZhang}. This is distinct from the C-ERM surrogate, which is trained via constrained optimisation to produce the max and min bounds over the version space.

\paragraph{LoRA configuration.}
We use Low-Rank Adaptation (LoRA) \cite{hu2021lora} to efficiently fine-tune the surrogate model in this tryout. The configuration is:
\begin{itemize}
    \item \textbf{Base model}: BERT-base-uncased (110M parameters)
    \item \textbf{LoRA rank} ($r$): 16 (low-rank dimension)
    \item \textbf{LoRA alpha} ($\alpha$): 32 (scaling parameter, $\alpha = 2r$)
    \item \textbf{LoRA dropout}: 0.1
    \item \textbf{Target modules}: \texttt{query} and \texttt{value} projections in attention layers
    \item \textbf{Trainable parameters}: $\sim$1.2M (1.1\% of base model)
\end{itemize}
This configuration reduces memory usage by $\sim$90\% compared to full fine-tuning while maintaining model capacity.

\paragraph{Surrogate metrics.}
We evaluate surrogate mimic behaviour using the following metrics:
\begin{itemize}
    \item \textbf{MSE}: Mean squared error between surrogate predictions $h(x)$ and black-box scores $h^\star(x)$ on held-out data
    \item \textbf{Rank correlation}: Spearman/Pearson correlation of surrogate vs black-box score rankings
    \item \textbf{Constraint satisfaction}: Fraction of queries where $|h(x) - h^\star(x)| \leq \lambda$ (with $\lambda=0.01$)
    \item \textbf{$\Delta$AUC gap}: Difference between surrogate-computed $\Delta$AUC and true black-box $\Delta$AUC
\end{itemize}

\begin{figure*}[t]
  \centering
  \includegraphics[width=\textwidth]{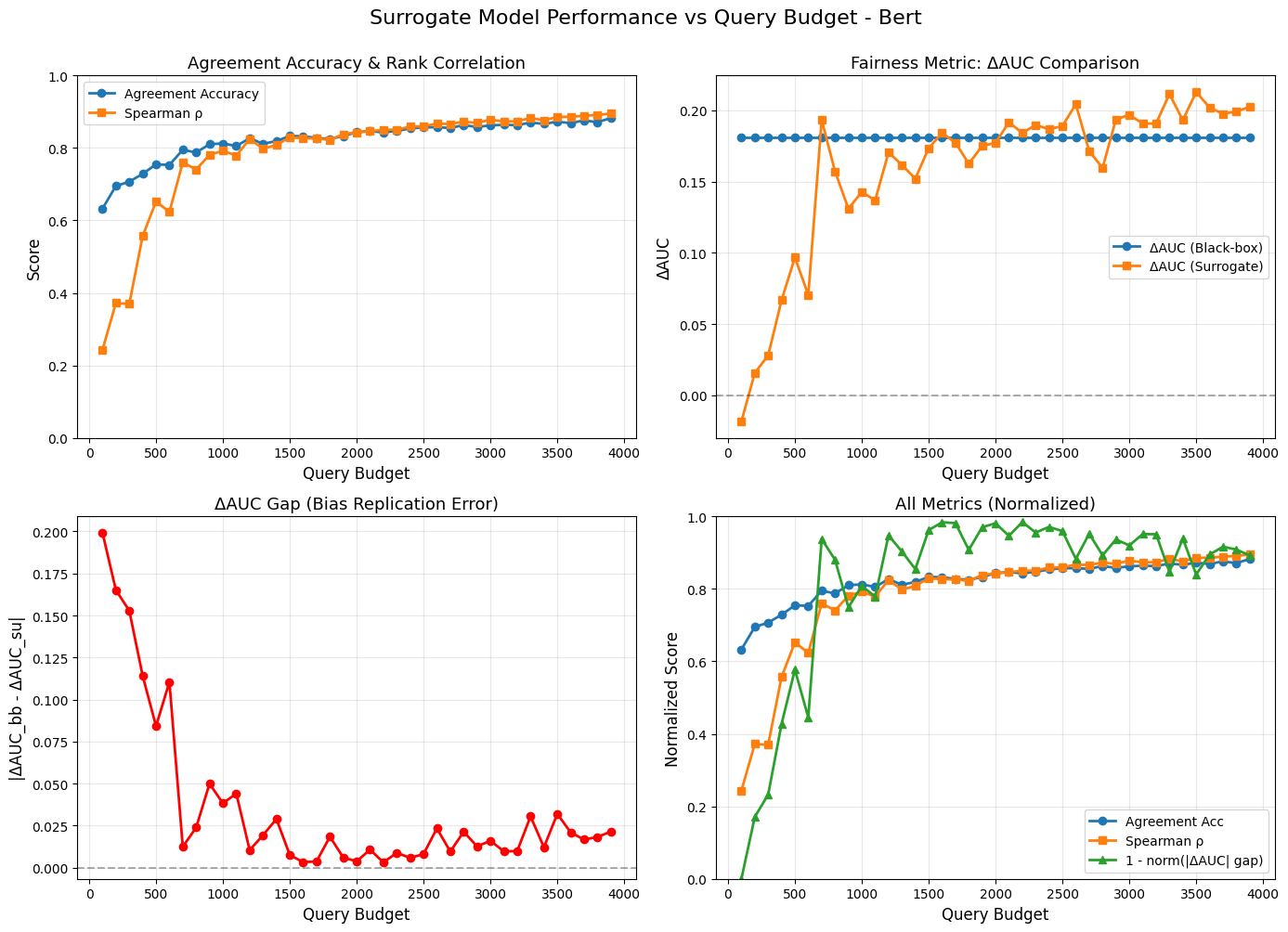}
  \caption{\textbf{Surrogate--black-box agreement vs.\ query budget.} 
  As the queried set grows, we report (i) pointwise agreement accuracy and Spearman rank correlation between surrogate scores $h(x)$ and black-box scores $h^\star(x)$, and (ii) the induced disparity replication error 
  $\bigl|\Delta\mathrm{AUC}(h^\star)-\Delta\mathrm{AUC}(h)\bigr|$. 
  While agreement and rank correlation increase steadily, the $\Delta\mathrm{AUC}$ replication error becomes small and stable only after roughly $500$--$750$ queries, indicating that many queries are required before the surrogate matches the black-box score geometry relevant for groupwise ranking.}
  \label{fig:surrogate_fidelity_bert}
\end{figure*}

\paragraph{Training procedure.}
The surrogate is trained on the current query set $S_t$ using a combined loss:
\[
\mathcal{L} = 0.2 \cdot \mathcal{L}_{\text{MSE}} + 0.8 \cdot \mathcal{L}_{\text{rank}},
\]
where $\mathcal{L}_{\text{MSE}}$ is mean squared error between surrogate probabilities and black-box scores, and $\mathcal{L}_{\text{rank}}$ is a margin ranking loss that preserves pairwise score orderings. Training uses AdamW optimizer with learning rate $5 \times 10^{-4}$, batch size 16, and 4 epochs per iteration.
\paragraph{Surrogate--black-box agreement and implications for constraint-based auditing.}
Figure~\ref{fig:surrogate_fidelity_bert} demonstrates how quick a BERT-based LoRA-surrogate (results are very similar with HateBERT as a surrogate) approaches the audited system as the query budget grows. Pointwise score agreement and rank correlation increase steadily and reach high values after a few hundred queries, but accurately reproducing the audit target requires substantially more supervision. In both settings (BERT and HateBERT), the surrogate’s induced disparity estimate $\Delta\mathrm{AUC}(h)$ aligns quantitatively with the black-box disparity $\Delta\mathrm{AUC}(h^\star)$ only after roughly $500$--$750$ queried examples. Before this query intervall, the surrogate often captures the correct direction of the disparity but exhibits large magnitude error in $\bigl|\Delta\mathrm{AUC}(h^\star)-\Delta\mathrm{AUC}(h)\bigr|$, indicating that matching scores in an average sense is not sufficient to match the groupwise ranking geometry that determines $\Delta\mathrm{AUC}$.

This gap matters for approaches that impose surrogate-based constraints in C-ERM, e.g., methods in the spirit of \cite{TomZhang} that treat $h(x)$ as a proxy for $h^\star(x)$ inside the constraint set. In our setting, reaching the regime where surrogate-based constraints would be reliable already consumes a significant fraction of the overall query budget, weakening the case for query-efficient third-party auditing. We therefore avoid using a learned surrogate as a constraint proxy in the certificate computation: BAFA’s certificate interval is derived by constrained optimisation using queried black-box scores only, not surrogate predictions as \cite{TomZhang} are doing. Learned representations are used only as auxiliary signals in the selection module (e.g., diversity-aware selection and BO features). Finally, to reflect realistic audit conditions for ranking-based metrics such as $\Delta\mathrm{AUC}$, we adopt a top-$k$ selection procedure that leverages known ground-truth labels for evaluation, rather than relying on surrogate-imputed scores. Together, these design choices keep BAFA effective in the low- to mid-budget regime where surrogate-based constraints are not yet dependable as visibly.

\end{document}